\documentclass[twoside]{article}

\usepackage[accepted]{aistats2022}

\usepackage{hyperref} %
\usepackage{url}
\usepackage{amsmath}
\usepackage{amsthm}
\usepackage{amssymb}
\usepackage[pdftex]{graphicx}
\usepackage{caption}
\usepackage{subcaption}
\usepackage{xcolor}         %
\usepackage{bm}
\usepackage{amsmath}
\usepackage{amsfonts}
\usepackage{color}
\usepackage[makeroom]{cancel}
\usepackage{cancel}
\usepackage{tabu}
\usepackage[normalem]{ulem} %
\usepackage{framed}
\usepackage{wrapfig}
\usepackage{amssymb}
\usepackage{pifont}
\usepackage{multirow}
\usepackage{booktabs}
\usepackage[flushleft]{threeparttable}
\usepackage[noabbrev,capitalise]{cleveref}
\usepackage{algorithm,algorithmic}

\usepackage{natbib}

\setcitestyle{authoryear, open={(}, close={)}}

\DeclareMathOperator*{\argmax}{arg\,max}

\begin{document}

\twocolumn[
\runningauthor{Bergamin, Mattei, Havtorn, Senetaire, Schmutz, Maaløe, Hauberg, Frellsen}
\runningtitle{Model-agnostic out-of-distribution detection using combined statistical tests}
\aistatstitle{Model-agnostic out-of-distribution detection\\ using combined statistical tests}
\aistatsauthor{Federico Bergamin\textsuperscript{*,1}, Pierre-Alexandre Mattei\textsuperscript{*,2}, Jakob D. Havtorn\textsuperscript{1,3}, Hugo Senetaire\textsuperscript{1} }

\aistatsauthor{Hugo Schmutz\textsuperscript{2,4}, Lars Maaløe\textsuperscript{1,3}, Søren Hauberg\textsuperscript{1}, Jes Frellsen\textsuperscript{1}}

\aistatsaddress{\textsuperscript{1}Technical University of Denmark \hspace{0.18cm} \textsuperscript{2}Université Côte d’Azur, Inria, LJAD, CNRS  \hspace{0.18cm}  \textsuperscript{3}Corti AI \hspace{0.18cm} \textsuperscript{4}TIRO, CEA}

]

\begin{abstract}
  We present simple methods for out-of-distribution detection using a trained generative model. These techniques, based on classical statistical tests, are model-agnostic in the sense that they can be applied to any differentiable generative model. The idea is to combine a classical parametric test (Rao's score test) with the recently introduced typicality test. These two test statistics are both theoretically well-founded and exploit different sources of information based on the likelihood for the typicality test and its gradient for the score test. We show that combining them using Fisher's method overall leads to a more accurate out-of-distribution test. We also discuss the benefits of casting out-of-distribution detection as a statistical testing problem, noting in particular that false positive rate control can be valuable for practical out-of-distribution detection. Despite their simplicity and generality, these methods can be competitive with model-specific out-of-distribution detection algorithms without any assumptions on the out-distribution. %
\end{abstract}

\section{Introduction}

The ability to recognise when data are anomalous, i.e.\@ if they originate from a distribution different from that of the training data, is a necessary property for machine learning models for safe and reliable applications in the real world. Historically, \citet{bishop1994novelty} proposed to use a one-sided threshold on the log-likelihoods of a learned model as a decision rule to identify outliers in a dataset. However, recently, \citet{nalisnick2018deep, hendrycks2018deep} showed that state-of-the-art deep generative models (DGMs) failed in this task, assigning higher a likelihood to out-of-distribution (OOD) data than in-distribution data. Most of the recent works focused on proposing new test statistics to alleviate the problem of using the plain likelihood, see \cref{sec:related_works} for details.

We believe that OOD detection should be formulated as statistical hypothesis testing \citep{nalisnick2019detecting, ijcai2021-292, haroush2021statistical}. Since the power of a single test depends on the out-distribution \citep{zhang2021understanding}, we propose to approach this problem by using a combination of multiple statistical tests. %
While the power of the combined test also depends on the out-distribution, we hypothesise that the combined test empirically will perform better, especially in situations where one of the statistics fails.
Furthermore, the use of the statistical testing framework has several advantages. Since we obtain a $p$-value, it is more natural deciding on a threshold as this corresponds to the significance level. In addition to that, it also allow us to correct for the multiple comparisons problem when identifying outliers in a dataset by controlling the number of Type I errors through the false discovery rate (FDR).

In summary, our contributions are the following:
\begin{itemize}
    \item We illustrate the benefits of combining multiple statistical tests to perform OOD detection with DGMs using well-established methods. This allows for a proper decision procedure to control the FDR in a real outlier detection setting.
    \item We revisit some proposed detection scores and highlight their alternative formulation as classical significance tests.
    \item Empirically we show the complementarity of the typicality and the score statistics and that their combination leads to a robust score for anomaly detection. %
\end{itemize}

\section{Using statistical tests for out-of-distribution detection}

We consider some data of interest that live in a space $\mathcal{X}$. Assume that we have a curated dataset $x_1,\ldots,x_m$, i.e.\@ there are no outliers, and we are interested in understanding if some new data $\tilde{x}_1,\ldots,\tilde{x}_n$ are collectively anomalies. In other words, we wonder whether or not $\tilde{x}_1,\ldots,\tilde{x}_n$ are likely to come from the same distribution that generated our curated dataset. We present in this section two different approaches for doing out-of-distribution detection using statistical tests: one based on classical parametric tests and one based on maximum mean discrepancy. A convenient property of the tests we consider is that they are all one-sided, which means we can expect them to be larger when the data are more likely to be OOD. This allows us to compute $p$-values by simply using the empirical CDF, which is hyperparameter-free.

Note that in this problem formulation, the case $n=1$ corresponds to the situation where we need to decide if a \emph{single} data point is out-of-distribution. This hardest setting will be of particular interest, and this is also the main focus of recent work, see \cref{sec:related_works}.

\subsection{Parametric tests for out-of-distribution detection}\label{sec:parametric}
The typical approach is to consider a parametric family $(p_\theta)_{\theta \in  \Theta}$ of probability densities over $\mathcal{X}$ and learn a suitable $\theta_0 \in  \Theta$ using any inference technique, for example maximum likelihood, and the clean data $x_1,\ldots,x_m$. Depending on the input domain, $(p_\theta)_{\theta \in  \Theta}$ could be composed of DGMs (in that case, $\theta$ would be neural network weights) or Gaussian mixture models (in that case, $\theta$ would be composed of means, covariances, and proportions). The question we wish to answer may then be phrased: \emph{is $p_{\theta_0}$ an appropriate model for $\tilde{x}_1,\ldots,\tilde{x}_n$?}

We choose to formalize this problem as a \emph{parametric test} whose alternative hypothesis is that $\tilde{x}$ is \emph{out-of-distribution}. More specifically, if we assume that $\tilde{x}_1,\ldots,\tilde{x}_n \sim_{\textup{i.i.d.}} p_{\tilde{\theta}}$ for some unknown $\tilde{\theta} \in \Theta$, we wish to test $\mathcal{H}_0 : \tilde{\theta} = \theta_0$ against $\mathcal{H} : \tilde{\theta} \neq \theta_0$, where the alternative hypothesis $\mathcal{H}$ is that the test points are OOD.

Many tests have been proposed for this purpose. The three most famous are the \emph{likelihood ratio test} of \citet{neyman1928use}, Rao's \citeyearpar{rao1948large} \emph{score test},
and \emph{the Wald test} \citep{wald1943tests}. These three classics are nicely reviewed by \citet{buse1982likelihood} or by \citet{rao2005score}, who called them the ``Holy Trinity''. A recent and interesting one is the \emph{gradient test} of \citet{terrell2002gradient}, which is reviewed in great detail in Lemonte's \citeyearpar{lemonte2016gradient} monograph.

Let us review the statistics of these four tests:
\begin{itemize}
    \item likelihood ratio statistic is $S_{LR} = 2 ( \ell ( \hat{\theta}) -\ell ( \theta_0))$,
    \item Wald statistic is $S_{W} = (\hat{\theta} - \theta_0)^T I(\hat{\theta}) (\hat{\theta} - \theta_0)$,
    \item score statistic is $S_{S} = \nabla \ell ( \theta_0)^T I(\theta_0)^{-1}  \nabla \ell ( \theta_0)$,
    \item gradient statistic is $S_{G} = \nabla \ell ( \theta_0)^T (\hat{\theta} - \theta_0)$,
\end{itemize}
where $ \ell ( \theta) = \log p_\theta (\tilde{x}_1,\ldots,\tilde{x}_n)$ is the likelihood function, 
$I({\theta}) = \mathbb{E}_{p_\theta}[\nabla \ell ( \theta)\nabla \ell ( \theta)^T]$ is the Fisher information matrix (FIM), and $ \hat{\theta} \in \argmax_{\theta \in \Theta} \ell ( \theta)$.

The likelihood ratio statistic, the Wald statistic and the gradient statistic all require to fit a model on the additional datapoints $\tilde{x}_1,\ldots,\tilde{x}_n$ in order to compute either $\ell(\hat{\theta})$ or $\hat{\theta}$. 
In our setting, if we want to use one of those statistics as an OOD score for a single example, we should fit a DGM on that single datapoint. \cite{xiao2020likelihood} did this for a variational autoencoder (VAE, \citealp{kingma2013auto,rezende2014stochastic}) by only re-fitting inference network (or encoder) to the additional example, which is a typical approach to dealing with out-of-sample data in VAEs, as argued by \cite{cremer2018inference} and \citet{mattei2018refit}.
However, much of the recent works in the literature \citep{ren2019likelihood, schirrmeister2020understanding, serra2019input} mainly focus on deriving different versions of what they call a likelihood ratio statistic.

We tried to derive a general way to compute both the Wald statistic and the gradient statistic, by computing $ \hat{\theta}$ with a few steps of a gradient-based optimization algorithm initialized at $ \theta_0$, but this resulted in a very unstable update leading to computational issues (results not shown). Therefore, in this work we focus on studying the relevance of the score statistic for performing out-of-distribution detection since it is the only statistic that does not require fitting an additional model to the OOD data.

\subsection{Maximum mean discrepancy for out-of-distribution detection}\label{sec:MMD}
Another way of approaching out-of-distribution detection from a testing perspective is through a \emph{two-sample test}. Denoting $p_\textup{data}$ the true training data distribution, the goal is to test $\mathcal{H}_0: \tilde{x}_1,\ldots,\tilde{x}_n \sim p_\textup{data}$ against $\mathcal{H}: \tilde{x}_1,\ldots,\tilde{x}_n \not\sim p_\textup{data}$, where the alternative hypothesis $\mathcal{H}$ again is that the test points are OOD.

A popular way of building statistics for two-sample tests is to use a measure of distance between $p_\textup{data}$ and the distribution of $\tilde{x}_1,\ldots,\tilde{x}_n$. The key idea here will be to use the trained generative model to build this measure of distance. To this end, we will use the \emph{maximum mean discrepancy (MMD)} of \citet{gretton2012kernel}, which is a kernel-based measure of distance. Then, $p_\theta$ will be used to specify an appropriate kernel.

More specifically, given a kernel whose feature map is $\Phi : \mathcal{X} \to \mathcal{H}$, the MMD between two distributions $P$ and $Q$ over $\mathcal{X}$ is defined as
\begin{equation}
	\textup{MMD}_\Phi(P,Q) = \| E_{X\sim P}[\Phi(X)] - E_{Y\sim Q}[\Phi(Y)] \|_\mathcal{H}. \hspace{-.1em}
\end{equation}
In our context, the test statistics will be of the form
\begin{multline}
	\label{eq:mmd}
	\textup{MMD}_\Phi\left(\frac{1}{m}\sum_{i=1}^{m} x_i,\frac{1}{n}\sum_{i=1}^n \tilde{x}_i\right) = \\\left\|\frac{1}{m}\sum_{i=1}^{m} \Phi(x_i) -  \frac{1}{n}\sum_{i=1}^{n} \Phi(\tilde{x}_i)\right\|_\mathcal{H},
\end{multline}
where $\Phi$ is a kernel feature map built using the generative model and $x_1,\ldots,x_m$ is the training data, i.e.\@ samples from  $p_\textup{data}$. When $\mathcal{H}$ is a simple finite-dimensional Hilbert space and $\Phi$ can be computed easily, then \eqref{eq:mmd} can be computed by going through the data and computing the means in an online fashion.%

As always with kernel methods, a key question is how to choose the kernel, or its feature map $\Phi$.
Here, we want to use the trained generative model $p_\theta$ to build our kernel feature map $\Phi$.

\paragraph{The Fisher kernel} An important example of kernel based on a generative model is the \emph{Fisher kernel} of \citet{jaakkola1999exploiting}. The embedding of this kernel is the Fisher score 
\begin{equation}
    \label{eq:Fisher_score}
	\Phi_\textup{Fisher} (x) = I(\theta)^{-\frac{1}{2}} \nabla \log p_\theta (x),
\end{equation}

and the corresponding reproducing kernel Hilbert space norm is just the $\ell_2$ norm: $|| \cdot || _\mathcal{H} = || \cdot ||_2$. In the case of the Fisher kernel, this means that \cref{eq:mmd} becomes:
\begin{multline}
	\label{eq:mmd_fisher}
	\textup{MMD}_{\Phi_\textup{Fisher}}\left(\frac{1}{m} \sum_{i=1}^{m} x_i,\frac{1}{n}\sum_{i=1}^n \tilde{x}_i\right) = \\ \left\|\frac{I(\theta)^{-\frac{1}{2}}}{m} \sum_{i=1}^{m}  \nabla \log p_\theta (x_i) - \frac{I(\theta)^{-\frac{1}{2}}}{n}  \sum_{i=1}^{n}  \nabla \log p_\theta (\tilde{x}_i)\right\|_2.
\end{multline}
We will see later that MMD with a Fisher kernel is closely related to the score statistic. In \cref{appendix:mahalanobis}, we additionally show that another popular OOD metric known as the \emph{Mahalanobis score} \citep{lee2018simple} can be interpreted as a MMD statistic with a certain Fisher kernel.

\paragraph{The typicality kernel} A very simple approach of embedding the data using $p_\theta$ is to choose $\Phi_\textup{Typical}(x) = \log p_\theta (x)$. Then, MMD is exactly equivalent to the \emph{typicality test statistic} of \citet{nalisnick2019detecting}, although this connection was not explicitly stated  by \citet{nalisnick2019detecting}. Because of this, we call the kernel $k(x,y) = \log p_\theta (x) \cdot \log p_\theta (y)$ the \emph{typicality kernel}.  %
While $\Phi_\textup{Typical}$ is not as well motivated as a kernel as $\Phi_\textup{Fisher}$, the concepts of typicality and typical set can be used to explain unintuitive behaviours of probability distributions in high-dimensional space as highlighted by \cite{nalisnick2018deep}. We also found that using this kernel generally gives good results for OOD tasks. An interesting analysis that we d not consider in this paper would be to study the properties of this kernel.

In general, neither of these two kernels are characteristic, meaning that our MMD can be zero even if the distributions are not identical. This could be solved by combining them with a characteristic kernel, as in \cite{liu2020learning}, at the price of including a new hyperparameter.

\section{Combining different test statistics}
\label{sec:combination_p_values}
For single-sample OOD detection, \citet{zhang2021understanding} proved that there is not a single statistic that is constantly better compared to all the possible alternatives of interest. For this reason, we believe that using a combination of different test statistics should lead to an overall better OOD detection in settings where a single statistic might fail. Assume we compute $k$ different test statistics $T_1,\dots, T_k$, each testing $\mathcal{H}_0$ against $\mathcal{H}$ as defined in \cref{sec:parametric,sec:MMD}. The goal is to combine these different tests into a single statistical test that ideally will perform better than the initial single tests. However, different tests can have different magnitudes and they can differ also in the direction of out-of-distribution detection, i.e.\@ for some statistics having a higher values is associated with being OOD, while for other smaller values are OOD. This makes a combination non-trivial.

\citet{morningstar2021density} proposed the density of states estimator (DoSE) to overcome this problem. They only focused on the single-sample detection task, i.e.\@ $n=1$ following our problem formulation. Their idea is to fit different nonparametric density estimators, such as a kernel-density estimator (KDE) or a one-class support vector machines (SVM), for each different statistic $T_1,\dots, T_k$ by using the values computed on the training set examples. For a single test example, $\tilde{x}_1$, they first compute $T_1,\dots, T_k$ and then combine those statistics by summing the different KDEs log-density. %
While this approach can be used for any type of statistic, and thus is more general, it uses less prior information. Indeed, if we use only statistics that are truly one-sided, then we assume that a method that leverages the true nature of the statistics should work better. In addition to that, fitting a KDE introduces an additional hyperparameter.

In our work, instead, we propose a different approach and leverage the fact that we use only one-sided test statistics. This setting is a well-studied problem in the literature both for independent \citep{fisher1925statistical, folks1971asymptotic} and dependent one-sided test statistics \citep{brown1975400, wilson2019harmonic}. All these approaches rely on the computation of $p$-values of each statistic for the test set $\tilde{x}_1,\ldots,\tilde{x}_n$. This corresponds to computing
$p_j = \textup{Pr}(T_j > t_j \mid \mathcal{H}_0)$, i.e.\@ the probability that the $j$'th test is bigger than the observed value under the null hypothesis $\mathcal{H}_0$, where we assume that each $T_j$ has a continuous distribution. Using $p$-values also solves the problem of the statistics having different scales. Indeed, $p$-values transform the different test statistics into the unit interval.

\paragraph{Computation of $p$-values} 
We want to approximate the distribution of the $p$-values $p_1, \ldots, p_k$ of $\tilde{x}_1,\ldots,\tilde{x}_n$ under the null hypothesis $\mathcal{H}_0$. When $\mathcal{H}_0$ is true, then $p_j$ is uniformly distributed on the interval $[0,1]$. To succeed in this, we should be able to compute $p_j = \textup{Pr}(T_j > t_j \mid \mathcal{H}_0)$, therefore we need to estimate the distribution of each statistic $T_j$ under $\mathcal{H}_0$. %
As done by \citet{nalisnick2019detecting}, we assume the existence of a validation set $\mathbf{X}'$ that was not used to train our generative model. From $\mathbf{X}'$ we bootstrap $S$ new datasets $\{\mathbf{X}^{'}_s\}_{s=1}^{S}$ of size $M'$ by using bootstrap resampling. When $n$ is small, for example $n=1$ or $n=2$, where $n=1$ corresponds to single-sample OOD detection, and the validation set is big, a convenient alternative to bootstrapping is to directly evaluate each test statistic $T_j$ on every single validation example. Asymptotically, this is equivalent to creating $S$ new datasets of size $M'=1$ when $S \rightarrow \infty$. In case of $n=2$, i.e.\@ two-samples OOD detection, and a big validation set we can simply bootstrap without resampling.
We then use these values to estimate the empirical distribution function (eCDF) of the considered statistic $T_j$ under $\mathcal{H}_0$. To obtain the $p$-values of test examples $\tilde{x}_1,\ldots, \tilde{x}_n$ for the test statistic $T_j = t_j$, we simply compute $p_j = 1 - \textup{Pr}(T_j < t_j \mid \mathcal{H}_0)$ using the eCDF.

\paragraph{Combining test statistics by combining $p$-values}
Fisher's \citeyearpar{fisher1925statistical} method is a procedure to combine different $p$-values $p_1,\dots, p_k$. This method assumes that all the considered test statistics are independent, and \citet{folks1971asymptotic} proved that it is asymptotically optimal among all methods of combining independent tests.
Given $T_1,\dots, T_k$ and corresponding $p$-values $p_1,\dots, p_k$, Fisher's method combines the $p$-values into a test statistic $X^2$ defined as
\begin{equation}
\label{eq:Fisher_method}
    X^2 \sim -2 \sum_{j=1}^{k} \ln (p_j).
\end{equation}
In case all null-hypotheses are accepted, the resulting test statistic $X^2$ follows a chi-squared distribution with $2k$ degrees of freedom. 
In the \cref{appendix:harmonic} , we also consider the Harmonic mean $p$-value \citep{wilson2019harmonic} as a way to combine $p$-values from different statistics. This method usually works best when the statistics are not independent.

\section{From test statistics to practical out-of-distribution scores}
\label{sec:section4}

Several of the test statistics that we consider make use of the inverse of the Fisher information matrix $I(\theta)$. The true Fisher information matrix requires an identifiable model to be invertible \citep{watanabe2009algebraic} and computing its inverse is $\mathcal{O}(m^3)$, where $m$ is the number of model parameters. For DGMs, the Fisher information matrix might not be invertible due to the fact that DGMs typically do not satisfy the identifiability condition. Also, the inversion may be computationally impractical, since state-of-the-art DGMs involve very high-dimensional parameter spaces $\Theta$. For the same reason, storing $I(\theta)$ can also be challenging.

We replace it by using a proxy matrix that has to be easy to compute and invert. A first idea is to simply replace $I(\theta)$ by the identity matrix. A more refined way is to look for a diagonal approximation. In \cref{appendix:cheap}, we describe cheap ways of computing such approximations. In particular, we will study two cases: the case where $I(\theta)$ is replaced by the identity matrix and the case where $I(\theta)$ is replaced by a diagonal matrix estimated using the training data.

A possible third option would be to estimate the diagonal of $I(\theta)$ using samples from the model. However, for autoregressive models as the PixelCNN, sampling is a sequential procedure and therefore it is computationally expensive to generate many samples when the input-space is high-dimensional. For this reason, we do not consider it in this work. More complex and precise approximations of the FIM exists, such as the Kronecker-factored Approximate Curvature (K-FAC, \citealp{martens2015optimizing}), but these are not defined for all types of layers used by state-of-the-art models.

\paragraph{On the difficulty of computing per-example gradients} Both the diagonal approximation of the FIM and the computation of the MMD with Fisher kernel of \cref{eq:mmd_fisher} require the gradient computation for all training and test examples. This is known as a costly procedure. For example, if we have to compute the gradient for $N$ examples using a simple fully connected network with $l$ layers of size $p$, the naive procedure of using a batch-size of dimension 1 is $\mathcal{O}(Nlp^2)$ \citep{goodfellow2015efficient}. While more efficient per-example gradient computations were proposed \citep{goodfellow2015efficient, rochette2019efficient}, these techniques can only be applied on simple fully connected or convolutional networks. While for this paper we relied on the naive solution of looping through every example one at the time, a more efficient solution is provided by the BackPACK library \citep{dangel2020backpack} which allows to compute the gradient with respect each sample in a minibatch. 

\subsection{Relationship between MMD with Fisher kernel and the score statistic and gradient norm} 
Depending on the choice of the Fisher information approximation, we can notice that there is a strong connection between the MMD using a Fisher kernel, the score statistic and the gradient norm in terms of expected OOD performance. Let us start by looking at the case where we approximate $I(\theta)$ with a diagonal matrix estimated using the training data. At the maximum likelihood estimate, we have that $\mathbb{E}[ \nabla \log p_\theta (x)] = 0$, i.e.\@ the first term inside the norm is $0$. Therefore, we expect that the differences between the OOD scores computed by using \cref{eq:mmd_fisher} will be preserved if we only consider $\left\|I(\theta)^{-1/2} \nabla \log p_\theta (\tilde{x}_1,\dots,\tilde{x}_n)\right\|_2$,
which corresponds to the square root of the score statistic. Since taking the square root still preserves the difference between values, we can expect that the MMD using a Fisher kernel will perform closely to the score statistic. The same reasoning also holds in case we replace the FIM with an identity matrix. In this specific case, instead, we will get that $\left\|{I}(\theta)^{-1/2} \nabla \log p_\theta (\tilde{x}_1,\dots,\tilde{x}_n)\right\|_2 = \left\|\nabla \log p_\theta (\tilde{x}_1,\dots,\tilde{x}_n)\right\|_2 $, which corresponds to considering the gradient norm. 

Computationally speaking, considering the score statistic instead of the MMD Fisher lets us avoid going through the entire training set to compute the average gradient (first term in \cref{eq:mmd_fisher}) while carrying the same information. Therefore, in this paper, we will mainly focus on the combination of the typicality test and the score statistic.

\subsection{Why does it make sense to combine the score statistic and the typicality test?}

Let us discuss our choice of combining the score statistic and the typicality test. We will try to look in which situations one of the test fails and the other works and vice versa. Both examples assume that the in-distribution data follows a $\mathcal{N}(0,I_D)$ distribution, and that the correct model has been learned by fitting $(\mathcal{N}(\theta,I_D))_{\theta \in \mathbb{R}^D}$ via maximum-likelihood. Even in this simple setting with no model misspecification, we will see that the two statistics that we consider may have very different strengths.

In this simple Gaussian case, the score statistic can be computed exactly and will be $||\tilde{x}_1 + \ldots + \tilde{x}_n ||_2^2 $. On the other hand, the typicality statistic will be $| (||\tilde{x}_1||_2^2 + \ldots + ||\tilde{x}_n||_2^2)/(2\cdot n) -  D/ 2| $. One interesting regime is the very high-dimensional one ($D \to \infty$). Indeed, by the law of large numbers, these random statistics become deterministic quantities.

\paragraph{Typicality fails, the score succeeds}

Assume that we have two independent OOD data samples that follow a product of truncated normal distributions, with density proportional to $$\mathcal{N}(x | 0,I_D) \cdot  \mathbf{1}\{x_1>0,\ldots,x_D>0\}.$$

We denote by $T_\text{score}^\text{ood},T_\text{score}^\text{id}$ and $T_\text{typicality}^\text{ood},T_\text{typicality}^\text{id}$ the statistics obtained when confronted with either OOD data from the truncated normal, or the in-distribution data. While these statistics are random in general, they will become deterministic when $D \to \infty$, by virtue of the law of large numbers.

For the typicality statistic, these two OOD samples will be indistinguishable from Gaussian ones. Indeed, when $D \to \infty$, both $T_\text{typicality}^\text{ood}$ and $T_\text{typicality}^\text{id}$ will be $\mathcal{O}(D)$.
On the other hand, for the score, one can show that
\begin{equation}
T_\text{score}^\text{ood} - T_\text{score}^\text{id} \sim 2 D \mu_\text{TN}^2,
\end{equation}
where $\mu_\text{TN}>0$ is the mean of the truncated normal distribution.

\paragraph{Typicality succeeds, the score fails}

Let us now consider as the OOD distribution a Dirac distribution with mean $0$. Suppose that we see a single sample from this distribution. In this case, the score statistic will be $0$, and will therefore not detect that the point is actually OOD. However, when $D$ is large, the typicality test will be able to declare that this point is anomalous, as shown by \citet{nalisnick2019detecting}.

Therefore, we have that the typicality test and the score statistic are complementary and measure a different type of information. In \cref{appendix:correlation}, we empirically show that they are not correlated, by plotting the two measures against each other and by computing the correlation matrix.

\section{Related works}
\label{sec:related_works}

Since \citet{nalisnick2018deep} and \cite{hendrycks2018deep}, different test statistics or methodologies for OOD detection using DGMs were proposed. Most of the recent solutions were highly influenced by three major lines of work: \emph{typicality set}, \emph{likelihood ratio} test statistics, and \emph{model misestimation}. 

The typicality set hypothesis was introduced by \citet{nalisnick2019detecting} as a possible explanation for the DGMs assigning higher likelihood to OOD data. The typicality set is the subset of the model full support where the model samples from and this does not intersect with the region of higher likelihood. While the typicality test was introduced for batch-OOD detection, \cite{morningstar2021density} shows that it also works well in the single-sample case. This is also confirmed by our own experiments. 

The likelihood ratio test statistic method by \citet{ren2019likelihood} assumes that every input is composed by a background component and a semantic component. For OOD detection, only the semantic component matters. In addition to a model trained on the in-distribution data, they proposed to train a background model on perturbed inputs data and then for each test example consider as OOD score the likelihood ratio between the two models. \cite{schirrmeister2020understanding}, instead, trained the background model on a more general distribution of images by considering 80 million general tiny images. Similarly to these approaches, \cite{serra2019input} argued that the failure of DGMs is due to the high-influence that the input complexity has on the likelihood. Therefore, they proposed to use a general lossless image compression algorithm as a background model.
All these methods, however, require additional knowledge of the OOD data for either choosing an image augmentation procedure to perturb the input data or for choosing a specific compressor. 

Another line of works blame the models themselves and not the test statistics. \cite{zhang2021understanding} argued that model misestimation is the main cause of higher likelihood assigned to OOD data. This can be due to both the model architecture and the maximum likelihood objective. \cite{kirichenko2020normalizing} and \cite{schirrmeister2020understanding} showed that normalizing flows can achieve better OOD performance despite achieving a worse likelihood if one changes some model design choices.
Other works in the literature focused on deriving specific test statistics that works only for a specific model, for example for VAEs \citep{xiao2020likelihood, maaloe2019biva, havtorn2021hierarchical}, or for normalizing flows \citep{kirichenko2020normalizing, ijcai2021-292}.

As mentioned in the introduction, we frame the OOD detection problem in terms of statistical tests problem. Recently, \cite{haroush2021statistical} showed that adopting hypothesis testing at the layer and channel level of a neural network can be used for OOD detection in the discriminative setting. They used both Fisher's method and Simes' method to combine class-conditional $p$-values computed for each convolutional and dense layer of a deep neural network. We focus on the unsupervised setting using DGMs and use hypothesis testing on statistics that can be computed on all differentiable DGM. As already explained in section \cref{sec:combination_p_values}, \cite{morningstar2021density} considered the combination of different statistics for OOD detection. The main difference with their approach is that we propose statistics that can be applied to any differentiable generative model and combine them by using Fisher's method, which takes advantage of using only one-sided independent statistics. Concurrently, \cite{choi2021robust} derived the score statistic by starting from the likelihood ratio statistic and applying a Laplace approximation. They computed the score statistic only for certain layers of the model and for a specific example, the OOD score is given by the infinity norm of these different layer scores after a ReLU operation. Our procedure differs both in the derivation of the score statistic and its usage since we compute the score statistic for the entire model.

\begin{figure*}[tbp]
    \includegraphics[width=\textwidth]{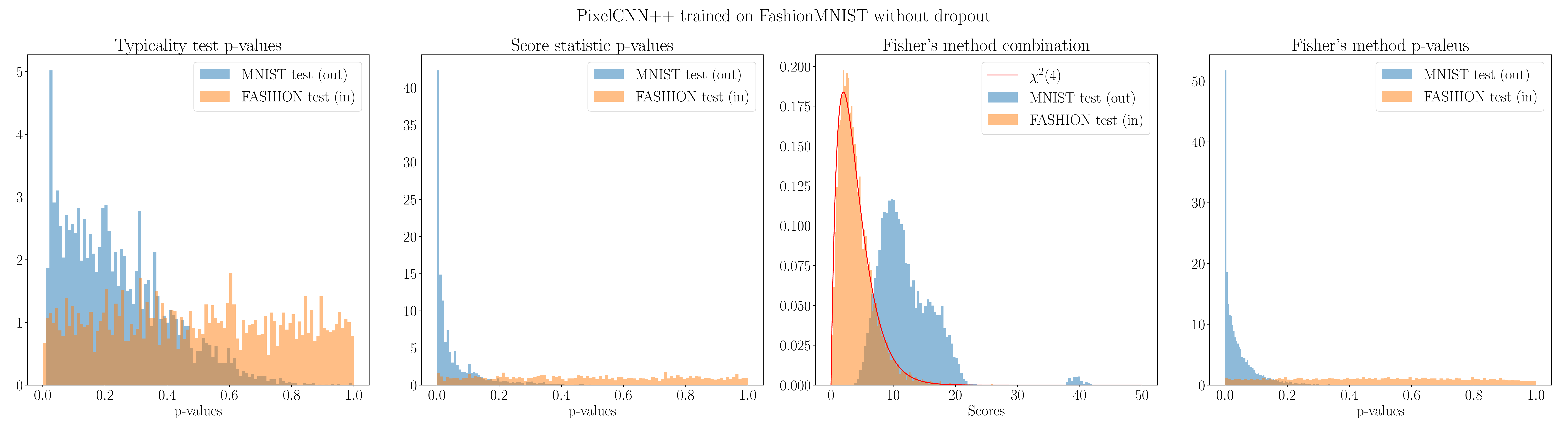}
    \caption{\emph{First plot}: $p$-values of the typicality test on the two test sets. We can see that under $\mathcal{H}_0$, they should be uniformly distributed. \emph{Second plot}: $p$-values of the score statistic. \emph{Third plot}: values obtained by the Fisher's method. In red, we plot the density function of a $\chi^2$-distribution with 4 degrees. This shows that the statistics are independent. \emph{Fourth plot}: $p$-values obtained of the combination. These plots refer to a PixelCNN++ trained on FashionMNIST without dropout.}
    \label{fig:comb_p_values}
\end{figure*} 

\begin{table*}[tb]
    \caption{AUROC$\uparrow$ for single-sample OOD detection. For Fisher's method we mean the combination of the typicality test and the test statistic. These are also combined using DoSE.}
    \resizebox{\textwidth}{!}{
        \scriptsize
        \begin{tabular}{ccccccc}
            \toprule
            &\multicolumn{6}{c}{\textsc{FashionMNIST (in) / MNIST (out)}}\\
            \cmidrule{2-7}
            & \multicolumn{4}{c}{\textcolor{blue}{\textsc{single statistics}}} & \multicolumn{2}{c}{\textcolor{red}{\textsc{combination}}}\\
            \cmidrule(r){2-5}\cmidrule(l){6-7}
            \textsc{models}  & \textcolor{blue}{\textsc{$\log p(x)$}} & \textcolor{blue}{\textsc{$\|\nabla \log p(x)\|_2$}} & \textcolor{blue}{\textsc{Typicality}} & \textcolor{blue}{\textsc{Score Stat}} & \textcolor{red}{\textsc{Fisher's method}} & \textcolor{red}{\textsc{DoSE$_{\textup{KDE}}$}} \\
            \midrule
            \textsc{PixelCNN++} (dropout)  &  0.0762 & 0.8709 & 0.8314 & \textbf{0.8822} & \textbf{0.9369}  & 0.8822 \\
            \textsc{PixelCNN++} (no dropout)  & 0.1048 &  \textbf{0.9532} &  0.7575 &  0.9381 & \textbf{0.9536} &  0.9382 \\
            \textsc{Glow} (RMSProp) &  0.1970 & 0.8904  & 0.4807 &  \textbf{0.9114} & 0.8598  &  \textbf{0.8901}\\
            \textsc{Glow} (Adam)  & 0.1223 & 0.7705 & 0.6987 &  \textbf{0.8745} & \textbf{0.8839}  & 0.8752 \\
            \textsc{HVAE}  & 0.2620 & 0.8714 &0.4884 & \textbf{0.9578} & 0.9383 & \textbf{0.9498} \\
            \bottomrule
            & & & & & & \\
            \toprule
            &\multicolumn{6}{c}{\textsc{CIFAR10 (in) / SVHN (out)}}\\
            \cmidrule{2-7}
            & \multicolumn{4}{c}{\textcolor{blue}{\textsc{single statistics}}} & \multicolumn{2}{c}{\textcolor{red}{\textsc{combination}}}\\
            \cmidrule(r){2-5}\cmidrule(l){6-7}
            \textsc{models}  & \textcolor{blue}{\textsc{$\log p(x)$}} & \textcolor{blue}{\textsc{$\|\nabla \log p(x)\|_2$}} & \textcolor{blue}{\textsc{Typicality}} & \textcolor{blue}{\textsc{Score Stat}} & \textcolor{red}{\textsc{Fisher's method}} & \textcolor{red}{\textsc{DoSE$_{\textup{KDE}}$}} \\
            \midrule
            \textsc{PixelCNN++} (model1)  & 0.1553 & \textbf{0.8006} &  0.6457 & 0.6407 & \textbf{0.6826} & 0.6571 \\
            \textsc{PixelCNN++} (model2) & 0.1567 & \textbf{0.7923} & 0.6498 &  0.7067 & \textbf{0.7300} & 0.7243 \\
            \textsc{Glow} (RMSProp)  &  0.0630 & 0.8585 & \textbf{0.8651} &  0.7940 &  \textbf{0.8683} & 0.8510 \\
            \textsc{Glow} (Adam)   &  0.0627 & 0.7844 &  \textbf{0.8624} &  0.7655 &  \textbf{0.8613} &  0.8588 \\
            \textsc{HVAE}  & 0.0636 & 0.8067  & \textbf{0.8679} & 0.7335 & \textbf{0.8603} & 0.8179 \\
            \bottomrule
            & & & & & & \\
            \toprule
            &\multicolumn{6}{c}{\textsc{CIFAR10 (in) / CIFAR100 (out)}}\\
            \cmidrule{2-7}
            & \multicolumn{4}{c}{\textcolor{blue}{\textsc{single statistics}}} & \multicolumn{2}{c}{\textcolor{red}{\textsc{combination}}}\\
            \cmidrule(r){2-5}\cmidrule(l){6-7}
            \textsc{models}  & \textcolor{blue}{\textsc{$\log p(x)$}} & \textcolor{blue}{\textsc{$\|\nabla \log p(x)\|_2$}} & \textcolor{blue}{\textsc{Typicality}} & \textcolor{blue}{\textsc{Score Stat}} & \textcolor{red}{\textsc{Fisher's method}} & \textcolor{red}{\textsc{DoSE$_{\textup{KDE}}$}} \\
            \midrule
            \textsc{PixelCNN++} (model1)  & 0.5153 & 0.5306 & \textbf{0.5458} & 0.5362 & \textbf{0.5563}  & 0.5477\\
            \textsc{PixelCNN++} (model2) & 0.5150 &  0.5230 & \textbf{0.5455} & 0.5325  & \textbf{0.5543} & 0.5453 \\
            \textsc{Glow} (RMSProp)  &  0.5206 & 0.5547 & 0.5507 & 0.\textbf{5801} & \textbf{0.5844} &  \textbf{0.5842} \\
            \textsc{Glow} (Adam)   & 0.5206 & 0.5593 & 0.5508  & \textbf{0.5692} & \textbf{0.5775} &  \textbf{0.5767} \\
            \textsc{HVAE}  & 0.5340 & 0.5280 &  0.5493 &  \textbf{0.5798} &  0.5879 &  \textbf{0.5941}\\
            \bottomrule 
        \end{tabular}
        \label{tab:single_sample_results}
    }
    \vspace*{-\baselineskip}
\end{table*}

\section{Experimental Setup}

To evaluate the performance of the combination of the typicality test and the score statistic in detecting OOD data, we follow the experiments of \citet{nalisnick2018deep, hendrycks2018deep} and considered the OOD detection task on three image dataset pairs that have been proven challenging for DGMs, i.e.\@ FashionMNIST \citep{xiao2017fashion} vs MNIST \citep{lecun1998mnist}, CIFAR10 \citep{krizhevsky2009learning} vs SVHN \citep{netzer2011reading}, and CIFAR10 vs CIFAR100. \cite{winkens2020contrastive} divide these tasks into \textit{far}-OOD tasks, where the in-distribution and out-distribution are different such as in the case of CIFAR10 against SVHN, and \textit{near}-OOD where the two distributions are pretty similar, such as CIFAR10 and CIFAR100. \textit{Near}-OOD tasks are usually most challenging.

For each task, we trained three different state-of-the-art DGMs, a PixelCNN++ \citep{salimans2017pixelcnn++}, a Glow model \citep{kingma2018glow}, and a hierarchical variational autoencoder \citep{kingma2013auto,rezende2014stochastic} with bottom-up inference (HVAE, \citealp{burda2015importance}). These are DGMs parametrized by neural networks that make different assumptions in the modelling choice of the target distribution. In addition to that, for PixelCNN++ and Glow we have a tractable likelihood while for HVAE we can only estimate a lower bound. A more in-depth description of these methods and additional results testing MNIST against FashionMNIST and SVHN against CIFAR10 can be found in \cref{appendix:mnist_svhn}. We also extensively analyzed, focusing mostly in the influence of the preprocessing, the results on CIFAR10 vs CelebA \citep{liu2015faceattributes} in \cref{appendix:celeba}. In \cref{appendix:gmm_ppca}, we also considered a Gaussian Mixture Model and a Probabilistic PCA as simple generative models.

\paragraph{Models} To analyze the effect of model architecture choices and optimization choice, we also consider different versions of the same model that reaches a similar log-likelihood. We consider 5 different models for each dataset pair. On FashionMNIST, we consider two Glow models, one trained using Adam and one using RMSProp and two PixelCNN++, trained with and without dropout. For CIFAR10, we consider two different PixelCNN++, one trained by us (model1) and one using a checkpoint given by the repository we used\footnote{ \href{https://github.com/pclucas14/pixel-cnn-pp}{\texttt{https://github.com/pclucas14/pixel-cnn-pp}}} (model2), and two Glow models (Adam and RMSProp). For both datasets, instead, we consider only one HVAE.

\paragraph{Baselines} We are mostly interested in testing our methods with other model-agnostic test statistics in the literature. Apart from using the plain likelihood as an OOD score, the only test statistic we are aware of that can be applied to any generative model without requiring any background model or OOD assumptions is the typicality test statistic of \cite{nalisnick2019detecting}. We also considered the gradient norm, which in general seem to work well but fails in the case of SVHN vs CIFAR10 (see \cref{appendix:mnist_svhn}). In addition to that, we compare our methods to a model-agnostic version of DoSE by \cite{morningstar2021density}, where we used KDEs to combine the score statistic and the typicality test statistic.

\paragraph{Evaluation} We compare our methods with the baselines by computing the area under the receiver operating characteristic curve (AUROC) as done in previous works \citep{hendrycks2018deep, ren2019likelihood, morningstar2021density}. We also evaluate our methods in terms of False Discovery Rate (FDR) control \cite{benjamini1995controlling}, i.e.\@ the proportion of false positive among the rejected hypothesis. Note that both quantities need to know the true label (OOD or in-distribution) to be computed.

\section{Results}

\paragraph{One-sample OOD}
We first evaluate our proposed method in the single-sample OOD detection task. Results are summarized in \cref{tab:single_sample_results}. We start by considering the OOD task on FashionMNIST against MNIST. Looking at the single statistics, we notice that the score statistic is the one that works the best and the combination of the typicality test and the score statistic usually improve the AUROC than the two standalone statistics. In addition to that, it is better than the combination of the two statistics by using a KDE. DoSE seems to perform better on Glow trained with RMSProp, where the typicality is failing. 

On natural images, instead, we have a different trend. The typicality test is better than the score statistic overall. The gradient norm surprisingly performs well in the two dataset pairs, but it fails badly when the model is trained on SVHN (see \cref{appendix:mnist_svhn}). Regarding the combination of the two statistics, the Fisher's method is always better than DoSE, but in this setting, it improves over the best of the single statistics three out of five times.
In the \textit{near}-OOD task, we have that both our method and DoSE using our suggested statistics perform closely. We want to highlight that for this challenging task we get results that are comparable with those reported in \cite{morningstar2021density}, but by using two model-agnostic statistics instead of three model-specific ones.
It can be noticed that the way we train our models has a strong influence on both the typicality test and the score statistic, although the models get the same test log-likelihood. In \cref{appendix:checkpoints}, we also show that this can happen between different checkpoints of the same model.

In \cref{fig:comb_p_values}, we show that the $p$-values distributions for both the typicality and the score statistic are uniformly distributed under the null-hypothesis and that the combination under the null follows a $\chi^2$ distribution with 4 degrees of freedom. This also supports the fact that the typicality test and the score statistic are independent.

\paragraph{Two-sample OOD}
As \citet{nalisnick2019detecting}, we consider how these test statistics change when performing two-sample OOD detection. Results are summarized in \cref{tab:two_sample_results}. As shown by \citet{nalisnick2019detecting}, the typicality improves but also the score statistic gets better if we consider more samples. Combining those leads to an improvement of performance in terms of AUROC with almost all the models. When training on FashionMNIST, the model can almost perfectly distinguish between the in-distribution test set and the OOD test set. While the performance improves for the two \textit{far}-OOD task, we have that the improvement is slightly less evident in the \textit{near}-OOD task of CIFAR10 vs CIFAR100.

\begin{table}[t]
\scriptsize
\caption{AUROC$\uparrow$ for two-sample OOD detection using the usual considered model.}\normalsize
    \resizebox{0.5\textwidth}{!}{
        \scriptsize
        \begin{tabular}{ccccc}
            \toprule
            &\multicolumn{4}{c}{\textsc{FashionMNIST (in) / MNIST (out)}}\\
            \cmidrule{2-5}
            \textsc{models}  & \textcolor{blue}{\textsc{Typicality}} & \textcolor{blue}{\textsc{Score Stat}} & \textcolor{red}{\textsc{Fisher's method}} & \textcolor{red}{\textsc{DoSE$_{\textup{KDE}}$}} \\
            \midrule
            \textsc{PCNN++} (drop.) &  0.9514 & 0.9828 & \textbf{0.9934} & \textbf{0.9912} \\
            \textsc{PCNN++} (no drop)  & 0.9081 &   0.9853 & \textbf{0.9916} &  \textbf{0.9921} \\
            \textsc{Glow} (RMSProp) & 0.6190 &  \textbf{0.9588} &  0.9187 & 0.7201  \\
            \textsc{Glow} (Adam)  & 0.8525 &  \textbf{0.9716} & \textbf{0.9708} & \textbf{0.9736} \\
            \textsc{HVAE}  & 0.6634 & \textbf{0.9881} & \textbf{0.9837} &  \textbf{0.9889}\\
            \bottomrule
            & & & &\\
            \toprule
            &\multicolumn{4}{c}{\textsc{CIFAR10 (in) / SVHN (out)}}\\
            \cmidrule{2-5}
            \textsc{models}  & \textcolor{blue}{\textsc{Typicality}} & \textcolor{blue}{\textsc{Score Stat}} & \textcolor{red}{\textsc{Fisher's method}} & \textcolor{red}{\textsc{DoSE$_{\textup{KDE}}$}} \\
            \midrule
            \textsc{PCNN++} (m1) & 0.7675 & 0.6555 & \textbf{0.7800} &  0.7046\\
            \textsc{PCNN++} (m2) & 0.7720 &   0.7235 & \textbf{0.8227} & 0.7850\\
            \textsc{Glow} (RMSProp)  & 0.9497 &  0.8624 & \textbf{0.9536} & 0.9379\\
            \textsc{Glow} (Adam) & 0.9480 &  0.8370 & \textbf{0.9519} & 0.9329\\
            \textsc{HVAE} &  \textbf{0.9623} &  0.7754 &  0.9560 & 0.9133\\
            \bottomrule
             & & & &\\
            \toprule
            &\multicolumn{4}{c}{\textsc{CIFAR10 (in) / CIFAR100 (out)}}\\
            \cmidrule{2-5}
            \textsc{models}  & \textcolor{blue}{\textsc{Typicality}} & \textcolor{blue}{\textsc{Score Stat}} & \textcolor{red}{\textsc{Fisher's method}} & \textcolor{red}{\textsc{DoSE$_{\textup{KDE}}$}} \\
            \midrule
            \textsc{PCNN++} (m1) &  0.5433 & 0.5450 &  \textbf{0.5540} &  \textbf{0.5508}\\
            \textsc{PCNN++} (m2) &  0.5435 &  0.5370 &  \textbf{0.5533} & 0.5470 \\
            \textsc{Glow} (RMSProp)  & 0.5550 &  \textbf{0.6211}  & 0.6165 &  \textbf{0.6233}\\
            \textsc{Glow} (Adam) & 0.5558  &   0.6073 & 0.6083 &   \textbf{0.6117}\\
            \textsc{HVAE} & 0.5594  & 0.6188  &  \textbf{0.6218} & \textbf{0.6273} \\
            \bottomrule
        \end{tabular}
        \label{tab:two_sample_results}
    }
    \vspace*{-\baselineskip}
\end{table}

\begin{figure}[tb]
    \centering
    \includegraphics[scale=0.47]{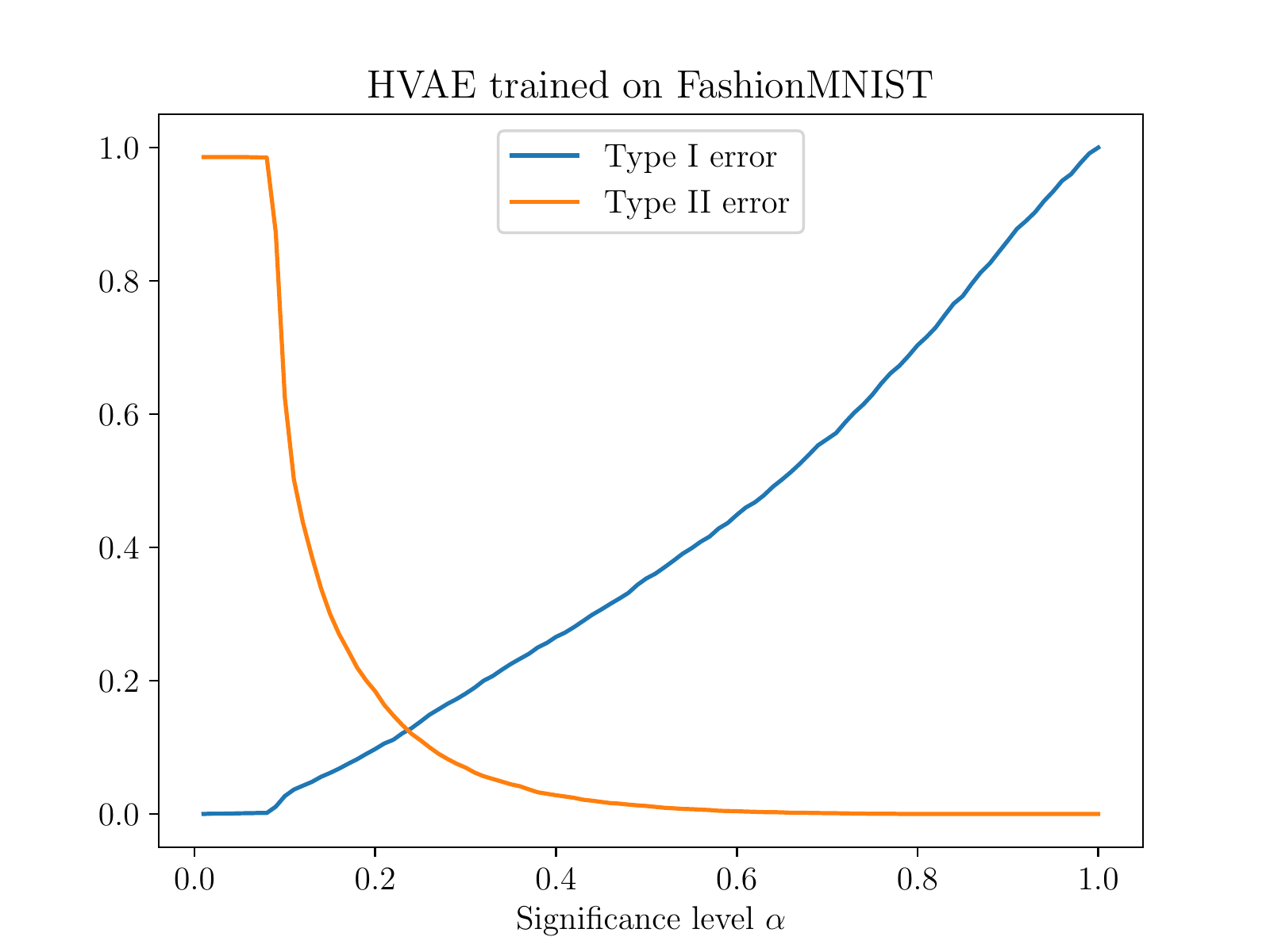}
    \caption{Type I (probability of an inlier to be classified as outlier) and Type II (probability of an outlier to be considered as inlier) errors versus the significance level $\alpha$ on the combination values. By using Benjamini-Hochberg correction, we get that the Type I error stays below identity line.}
    \label{fig:type1}
\end{figure}

\subsection{Practical OOD detection with FDR control}
One of the advantages of framing the problem as multiple testing is that we have a well-defined procedure to decide on which hypotheses to reject while controlling the False Discovery Rate (FDR, \citealp{benjamini1995controlling}). Imagine we are interested in finding the outliers from the dataset given by the combination of the two test-sets but we do not want to discard too many inliers, then we can use the Benjamini-Hochberg (BH) procedure \citep{benjamini1995controlling} to decide a threshold and reject all hypothesis below that threshold. For a specific significance level $\alpha$, the procedure guarantees that the FDR stays below that level. Therefore, we can guarantee that the rate of inliers that are classified as outliers is less than the chosen $\alpha$.

We leverage the fact that when the null hypothesis is true and the $p$-values are independent, then the scores obtained by combining $k$ different statistics are $\chi^2_{2k}$ distributed to compute the $p$-values. Alternatively, the procedure can be also applied to the $p$-values of a single test-statistic.
Usually, it is better to use a FDR control when it is actually possible to make few false discoveries, i.e.\@ when we have a strong statistic. Therefore, we expect the procedure to work well when the AUROC is good, for examples on models trained on FashionMNIST.

As can be seen in \cref{fig:type1}, we have that the Type I ratio line stays below the identity line, meaning that the BH correction is working. When deciding for a specific threshold $\alpha$, we usually have to trade-off between Type I and Type II error and in most cases the threshold to choose depends on the application domain. Ideally, we would like to have a low Type I and a low Type II error rate, meaning that we are not considering a lot of in-distribution examples as OOD and at the same time considering a lot of outliers as in-distribution. \cref{fig:type1} shows that we can achieve this for low values of $\alpha$. When training on CIFAR, instead, we are able to control the FDR only from a certain significance level (see \cref{appendix:BH_cifar}). This is expected given that the AUROC is not as good as when testing on MNIST.

\section{Discussion and Conclusions}
In this paper we studied the task of out-of-distribution detection using deep generative models and a combination of multiple statistical tests. We tested our method using different state-of-the art DGMs on classic image benchmark for OOD detection. We found that combining the two statistic leads to a more robust score that in some cases is close to state-of-the-art model-specific scores that require more assumptions. We also noticed that both the model design choice and the optimization choices have an influence on the score we are computing. 

When considering only one-sided independent statistics, we showed that the Fisher's method tends to works better than combine them by summing the log-density of a KDE. We also noticed that the score statistic tends to perform a bit worse when the number of parameters of the models increases, i.e.\@ in the context of natural images. One possible reason can be that in this setting the diagonal approximation is not good, and therefore one could consider different approximations, such as K-FAC. 

DGMs have recently been used for handling missing data (see e.g.\@ \citealp{mattei2019miwae, ma2018eddi, nazabal2020handling, ipsen2020not}). An interesting future direction would be to extend these OOD detection methods to handle missing values.

The methods  presented in this paper can also easily be applied when using model-specific one-sided statistics. In addition to obtain a more accurate score if one want to combine the test statistics, this also allows one to use well-defined procedure to control the FDR when choosing a which example to mark as outliers. Having this control, is necessary when we want to apply these methods in real settings.

\subsubsection*{Acknowledgements}
Federico Bergamin and Pierre-Alexandre Mattei contributed eqaully to this paper, which is indicated by the asterisk (*) in the author list. The work was supported by the Innovation Fund Denmark (0175-00014B and 0153-00167B), the Independent Research Fund Denmark (9131-00082B) and the Novo Nordisk Foundation (NNF20OC0062606 and NNF20OC0065611). Furthermore, it was supported by the French government, through the 3IA Côte d’Azur Investments in the Future project managed by the National Research Agency (ANR) with the reference number ANR-19-P3IA-0002.

\bibliography{biblio}

\clearpage
\appendix

\thispagestyle{empty}

\onecolumn \makesupplementtitle

\section{Crude approximation of the Fisher information}
\label{appendix:cheap}

The Fisher information is defined as:
\begin{equation}
    I({\theta}) =  \mathbb{E}_{x \sim p_\theta}[\nabla \log {p_\theta}(x) \nabla  \log {p_\theta}(x) ^T].
\end{equation}

A crude diagonal approximation can be computed by simply estimating the diagonal of $I({\theta})$ and setting all off-diagonal elements to zero. Such diagonal approximations have been used in machine learning for decades: for instance, \citet[Section 3.12.2]{lecun1987} used a similar approximation of the Hessian matrix, and called it ``outrageously simplifying". Much more complex approximations have been derived, although diagonal approximations have been consistently used (e.g. by \citealp{kirkpatrick2017overcoming}, who used essentially the same approximation in a supervised context), and are linked to several adaptive optimisation techniques like Adam \citep{kingma2014adam} or RMSProp \citep{tieleman2012lecture}. A good discussion on these issues is provided in Martens's \citeyearpar{martens2020new} recent review.

The approximation we used in the paper works as follows:
\begin{itemize}
    \item By using the training examples $x_1,...,x_T$, we form the estimate
    $$ {D}_T({\theta}) = \frac{1}{T} \sum_{t=1}^T \textup{Diag}(\nabla \log {p_\theta}(x_t)^2),  $$
    where the square in $\nabla \log p_\theta(x_t)^2$ is computed elementwise.
    \item While we could directly use ${D}_T({\theta})$ as an estimate. A slightly more refined approach is to slightly regularise $D_T({\theta})$. Following \citet{martens2020new}, our final estimate of the Fisher information matrix is
    \begin{equation}
    	\hat{I}_T({\theta}) = (D_T({\theta}) + \varepsilon)^\xi,
    \end{equation}
with all operations performed elementwise. The diagonal matrix $\hat{I}_T({\theta})$ is then easy to invert and can be used to compute our statistics.
\end{itemize}

\paragraph{How to choose $\varepsilon$ and $\xi$?} The Adam optimizer uses a similar estimate, with default hyperparameters $\varepsilon = 10^{-8}$ and $\xi = 1$. As argued by \citet{martens2020new}, it can be interesting to use $\xi <1$  in order to diminish the influence of extreme values of $D_T({\theta})$. In particular, \citet{martens2020new} suggests taking $\xi = 0.75$. When $\xi \longrightarrow 0$, then $\hat{I}_T({\theta})$ will approach the identity matrix. We tested the two settings by using a PixelCNN++ trained on CIFAR. Results are shown in table~\ref{tab:martens}. In terms of OOD detection, it seems that using $\varepsilon = 10^{-8}$ and $\xi = 1$ is slightly better. All results presented in the paper and in the supplementary material are computed by using $\varepsilon = 10^{-8}$ and $\xi = 1$.

\paragraph{A few notes on the computation of $D_T({\theta})$} While it seems more sensible to use samples $x_1,...,x_m\sim p_\theta $ from the model, we decided to simply reuse the training data $x_1,...,x_T $ instead. There are two computational advantages to this. The first one is that sampling many data points can be expensive (in particular for deep autoregressive models à la PixelCNN). The second advantage is that, if we wish to compute a MMD statistic, such as the MMD with the Fisher kernel or the MMD typicality (that require the average of gradient or the average log-likelihood over the training), computing the average of the square of the gradient costs very little. One can just do a single loop over the data, and use the usual formulas for online estimation of a mean, see \cref{algo_description}.

\paragraph{Do we really need to approximate the diagonal of $I({\theta})$?} 
Another possibility is to just use the identity matrix as FIM instead of of approximating the diagonal through the procedure explained above. In our experiments (see table~\ref{tab:mmd} and table~\ref{tab:mmd_pair2}), we can see that sometimes using the identity matrix seems to work equally well or a bit better for some models trained on FashionMNIST and CIFAR10. However, when we train on SVHN or MNIST, there are a cases where the statistic that is using the identity matrix as approximation fails, sometimes being worse than random chance. In those setting, using the diagonal approximation leads to way better results. Therefore, considering a test statistic that uses the diagonal approximation of the FIM is more robust for OOD detection.

\begin{table*}[tb]
    \centering
     \caption{AUROC$\uparrow$ for single-sample OOD detection. Comparison between two different estimates of the Fisher information matrix. For ($\ddagger$) we used the Adam parameter choice, i.e.\@ $\varepsilon = 10^{-8}$ and $\xi = 1$. For ($\mathsection$), instead, we used  $\varepsilon = 10^{-8}$ and $\xi = 0.75$, as suggested by \citet{martens2020new}. As a results we have that using Adam parameters choice is slightly better for our task.}
        \small
        \begin{tabular}{ccccc}
            \toprule
            &\multicolumn{4}{c}{\textsc{CIFAR10 (in) / SVHN (out)}}\\
            \cmidrule{2-5}
            \textsc{models}  & \textcolor{blue}{\textsc{MMD Diagonal}} &  \textcolor{blue}{\textsc{Typicality}} & \textcolor{blue}{\textsc{Score Stat}} & \textcolor{red}{\textsc{Fisher's method}} \\
            \midrule
            \textsc{PixelCNN++} (model2) ($\ddagger$) & 0.7070 & 0.6498  &  0.7067 & 0.7300\\
            \textsc{PixelCNN++} (model2) ($\mathsection$) & 0.6881 &  0.6498  & 0.6878 & 0.7176\\
            \bottomrule
            ($\ddagger$) With $\varepsilon = 10^{-8}$ and $\xi = 1$ \\
            \hspace{0.4cm}($\mathsection$) With $\varepsilon = 10^{-8}$ and $\xi = 0.75$
        \end{tabular}
        \label{tab:martens}
\end{table*}

\section{The Mahalanobis score as MMD}
\label{appendix:mahalanobis}

\citet{lee2018simple} introduced a simple metric to perform OOD detection with a trained deep classifier. The key idea is to train a simple generative model (linear discriminant analysis) in the feature space of the classifier. Let $y$ denote the labels, and $z = f(x)$ the data in feature space. In the simplest case, $f$ is just the trained deep net devoid of the last softmax layer. The linear discriminant analysis model is
\begin{equation}
y \sim \textup{Cat}(\pi), \; \; z | y \sim \mathcal{N}(\mu_y, \Sigma),
\end{equation}
where $\mu_1,...,\mu_K$ are class-dependent means, $\Sigma$ a common covariance matrix, and $\pi_1,...,\pi_K$ are the class proportions, estimated by maximum-likelihood. The \emph{Mahalanobis score} is then
\begin{equation}
M(x) = \max_{k \in \{1,...,K \}} - (z - \mu_k)^T \Sigma^{-1} (z - \mu_k),
\end{equation}
which may be rewritten
\begin{equation}
M(x) = \max_{k \in \{1,...,K \}} p(z|k),
\end{equation}
under the assumption of equal class proportions (i.e.\@ $\pi_1 = ... = \pi_K = 1/K$).

We show here that it is possible to re-interpret this score as a MMD score with a certain Fisher kernel. 
The generative model induced on $z$ by linear disciminant analysis is a Gaussian mixture:
\begin{equation}
p_{\pi, \mu , \Sigma }(z) = \sum_{k=1}^K \pi_k \mathcal{N}(z | \mu_k, \Sigma).
\end{equation}
If we want a powerful deep kernel, it seems somewhat natural to consider the Fisher kernel associated with this generative model. The most important part of this mixture model are arguably the class-specific means (indeed, the model has been trained to discriminate the classes as well as possible). Therefore, we will only include these means in the Fisher kernel, and look at
\begin{equation}
	\Phi_\textup{Fisher} (x) = I(\mu)^{-1/2} \nabla_\mu  \log p_{\pi, \mu , \Sigma }(z),
\end{equation}
assuming that $\pi$ and $\Sigma$ are fixed at their maximum likelihood estimates. Similar mixture-based Fisher kernels have been very popular in the past, and were actually a key element of state-of-the art classification models on Imagenet before deep nets won the competition \citep{perronnin2010improving}. Our idea is to re-use ideas introduced by this computer vision litterature. Under the assumption that the Gaussian clusters are well-separated, \citet{tanaka2013fisher}, extending an earlier analysis of \citet[Appendix A]{sanchez2013image}, showed that
\begin{equation}
	[\Phi_\textup{Fisher} (x)]_{\mu_k} \approx \sqrt{\frac{p(z|k)}{\pi_k }}  \Sigma^{-1/2} (z - \mu_k).
\end{equation}
Now, using the fact that the expected value of the score is approximatively zero, we can write that
\begin{equation}
\textup{MMD}_{\Phi_\textup{Fisher}}^2 \approx \sum_{k=1}^K  || [\Phi_\textup{Fisher} (x)]_{\mu_k} ||_2^2 \approx \sum_{k=1}^K \frac{p(z|k)}{\pi_k} (z - \mu_k)^T \Sigma^{-1} (z - \mu_k).
\end{equation}

Using again the fact that the clusters are well-separated, we may say that $z|k$ is approximatively a point mass at the most probable label,  i.e.\@ that $p(z|k) \approx \delta_k^{\text{argmax}_c p(z|c)}$. This leads to the approximation 
\begin{equation}
\textup{MMD}_{\Phi_\textup{Fisher}}^2 \approx \max_{k \in \{1,...,K\}} \frac{1}{\pi_k} (z - \mu_k)^T \Sigma^{-1} (z - \mu_k).
\end{equation}
Finally, assuming that the class proportions are equal leads to the equivalence of $\textup{MMD}_{\Phi_\textup{Fisher}}$ and the Mahalanobis score.

\section{More Information on the experimental setup}

\subsection{A bit more background}
The three considered DGMs are both parametrized by neural networks but they differ in the way they model the data distribution of interest. Assume we are interested in approximating a target distribution $p^*(\mathbf{x})$, for example a distribution of natural images, as it is done when using CIFAR10. PixelCNN++ is an autoregressive model and it models $p^*(\mathbf{x})$ as a product of conditional distribution over the variables, i.e.\@ $p(\mathbf{x}) = p(x_1) \prod_{d=2}^D p(x_d \mid \mathbf{x}_{<d})$, where $\mathbf{x}_{<d} = [x_1, \dots, x_{d-1}]^T$. Glow is a normalizing flow model and it approximate $p^*(\mathbf{x})$ by using a sequence of bijiective transformations starting from a simple distribution, also called base distribution. If we use only a single invertible transformation $f$, the normalizing  flow is defined as $\mathbf{x} = f(\mathbf{z})$, where $\mathbf{z} \sim p_{Z}(\mathbf{z})$, and $p_{X}(\mathbf{x}) = p_{Z}(\mathbf{z})|\det J_f(\mathbf{z}) |^{-1}$, where we used the change of variable formula. For these two types of model we have a tractable likelihood that can be used to optimize the model parameters. The Variational Autoencoder (VAE), instead, is a framework to model the data with a latent variable model, i.e.\@ $p(\mathbf{x}, \mathbf{z}) = p(\mathbf{x} \mid \mathbf{z}) p(\mathbf{z})$, where $\mathbf{x}$ is the observed input data and $\mathbf{z}$ is a stochastic latent variable and the prior distribution $p(\mathbf{z})$ is usually a standard Normal. Since the posterior $p(\mathbf{z}\mid\mathbf{x})$ is not tractable, a variational distribution $q_{\phi}(\mathbf{z}\mid\mathbf{x})$ is used as an approximation. Due to the intractability of the posterior, we cannot directly optimize the likelihood of the model, but instead the model parameters are optimized by maximizing the evidence lower bound (ELBO): $\log p_{\theta}(\mathbf{x}) \geq \mathbb{E}_{q_{\phi}(\mathbf{z}\mid \mathbf{x})} \left[ \log \frac{p(\mathbf{x}, \mathbf{z})}{q_{\phi}(\mathbf{z}\mid \mathbf{x})} \right] \equiv \mathcal{L} $. In this work we are considering an Hierarchical VAE (HVAE) with bottom-up inference as done in \cite{havtorn2021hierarchical}. This is an extension of the VAE framework that consider an hierarchy of $L$ latent variables $\mathbf{z} = \mathbf{z}_1,\dots, \mathbf{z}_L$. The bottom-up inference is defined as $q_{\phi}(\mathbf{z}\mid \mathbf{x}) = q_{\phi}(\mathbf{z}_1\mid \mathbf{x})\prod_{i=2}^L q_{\theta}(\mathbf{z}_i \mid \mathbf{z}_{i-1})$, while the generative path is top-down, meaning $p_{\theta}(\mathbf{x}\mid \mathbf{z}) = p(\mathbf{x}\mid \mathbf{z}_1)p_{\theta}(\mathbf{z}_1\mid \mathbf{z}_2)\cdots p_{\theta}(\mathbf{z}_{L-1}\mid \mathbf{z}_L)$. This is still trained by maximizing the ELBO. For a more in-depth explanation of these models we refer to their papers.

\subsection{Generative model details}
We will briefly describe the different model architectures and training procedures used in this paper. Since most of the models are taken from public code repositories and related papers, we will mostly invite the reader to have a look at the cited paper for a more in-depth description of the training details.
For MNIST, CIFAR10, and FashionMNIST we used 3000 examples from the test set as validation set. For SVHN, instead, we used 6032 datapoints from the test set as validation, leaving the remaining 20000 examples as test set. In Table~\ref{tab:log_like}, we reported test log-likelihood of the models used in this paper. 

\paragraph{PixelCNN++} For PixelCNN++ we used the code available in this repository\footnote{\href{https://github.com/pclucas14/pixel-cnn-pp}{\texttt{https://github.com/pclucas14/pixel-cnn-pp}}}. For the greyscale images, we used one residual block per stage with 32 filters and 5 logistic components in the discretized mixture of logistics. For natural images, instead, we used 5 residual blocks per stage with 160 filters and 10 components in the mixture. We trained all the models using Adam optimizer.

\paragraph{Glow} For training Glow models we follow \citet{kirichenko2020normalizing} and their repository\footnote{\href{https://github.com/PolinaKirichenko/flows_ood}{\texttt{https://github.com/PolinaKirichenko/flows\_ood}}}. They closely follow \citet{nalisnick2018deep} and \cite{kingma2018glow} implementation for multi-scale Glow, where a scale is defined as the sequence of actorm, invertible $1\times1$ convolution and coupling layers. While \citet{kirichenko2020normalizing} only considers the RMSProp optimizer, we trained two different models, one using RMSProp and one using Adam with batch-size 32. For the greyscale dataset our Glow is made up of 2 scales with 16 coupling layers, and a 3-layers highway network with 200 hidden units is used to predict the scale and shift parameters. For CIFAR10 and SVHN, instead, we used 3 scales with 8 coupling layers, and 400 hidden units for the 3-layers highway network. %
For a more in-depth description, we refer to the codebase and the Appendix C of \citet{kirichenko2020normalizing}.

\paragraph{Hierarchical VAE} We follow \cite{havtorn2021hierarchical} for both model architecture design and training choices for our hierarchical VAEs. We used their open-sourced repository\footnote{\href{https://github.com/JakobHavtorn/hvae-oodd}{\texttt{https://github.com/JakobHavtorn/hvae-oodd}}}. As mentioned in the paper, the HVAE model we used has a bottom-up inference path and a top-down generative path. We trained each model for 1000 epochs using Adam optimizer with learning rate $3e-4$ and a batch-size of $128$. All models were initialized using the data-dependent initialization and they used weight-normalization \citep{salimans2016weight}. In addition to that, we always consider a hierarchy of three latent variables. For greyscale images (MNIST and FashionMNIST) we used a latent dimension of $8-16-8$ for $\mathbf{z}_1, \mathbf{z}_2, \mathbf{z}_3$ respectively, while for natural images (CIFAR10 and SVHN) we used $8-16-32$. For a more in depth description of the model, we refer to Appendix B of \citet{havtorn2021hierarchical}.

\begin{table*}[tb]
    \caption{Test log-likelihood (bits/dim) achieved by the models used in the paper.}
    \begin{subtable}{.5\textwidth}
        \centering
        \scriptsize
        \begin{tabular}{cc}
            \toprule
            \multicolumn{2}{c}{\textsc{Models trained on FashionMNIST)}}\\
            \midrule
            \textsc{models}  & \textcolor{black}{\textsc{Log-Likelihood (bits/dim)}}\\
            \midrule
            \textsc{PixelCNN++} (dropout) & 2.75 \\
            \textsc{PixelCNN++} (no dropout) & 2.72 \\
            \textsc{Glow} (RMSProp) &  3.04\\
            \textsc{Glow} (Adam)  & 3.02 \\
            \textsc{HVAE} ($**$) & 0.43 \\
            \bottomrule
            &   \\
            \toprule
            \multicolumn{2}{c}{\textsc{Models trained on CIFAR10}}\\
            \midrule
            \textsc{models}  & \textcolor{black}{\textsc{Log-Likelihood (bits/dim)}}\\
            \midrule
        
            \textsc{PixelCNN++} (model1)  & 2.94  \\
            \textsc{PixelCNN++} (model2)  & 2.94 \\
            \textsc{Glow} (RMSProp)   & 3.62 \\
            \textsc{Glow} (Adam)  & 3.62  \\
            \textsc{HVAE}  & 3.87 \\
            \bottomrule
            ($**$) Binarized FashionMNIST
        \end{tabular}
        \end{subtable}
        \begin{subtable}{.5\textwidth}
        \centering
         \scriptsize
        \begin{tabular}{cc}
            \toprule
            \multicolumn{2}{c}{\textsc{Models trained on MNIST}}\\
            \midrule
            \textsc{models}  & \textcolor{black}{\textsc{Log-Likelihood (bits/dim)}}\\
            \midrule
            \textsc{PixelCNN++} (dropout) & 0.90 \\
            \textsc{Glow} (RMSProp)  &  1.32 \\
            \textsc{Glow} (Adam)   & 1.30 \\
            \textsc{HVAE} ($**$)   &  0.16\\
            \bottomrule
            &   \\
            \toprule
            \multicolumn{2}{c}{\textsc{Models trained on SVHN}}\\
            \midrule
            \textsc{models}  & \textcolor{black}{\textsc{Log-Likelihood (bits/dim)}}\\
            \midrule
            \textsc{PixelCNN++} (dropout)  & 1.58 \\
            \textsc{Glow} (RMSProp)   & 2.23 \\
            \textsc{Glow} (Adam)  & 2.21  \\
            \textsc{HVAE}  & 2.38 \\
            \bottomrule
            ($**$) Binarized MNIST
        \end{tabular}
        
        \end{subtable} 
        \label{tab:log_like}
\end{table*}

\section{Additional results}

\subsection{Typicality test and score statistic are uncorrelated}
\label{appendix:correlation}
\begin{figure}[tb]
    \centering
    \includegraphics[scale=0.5]{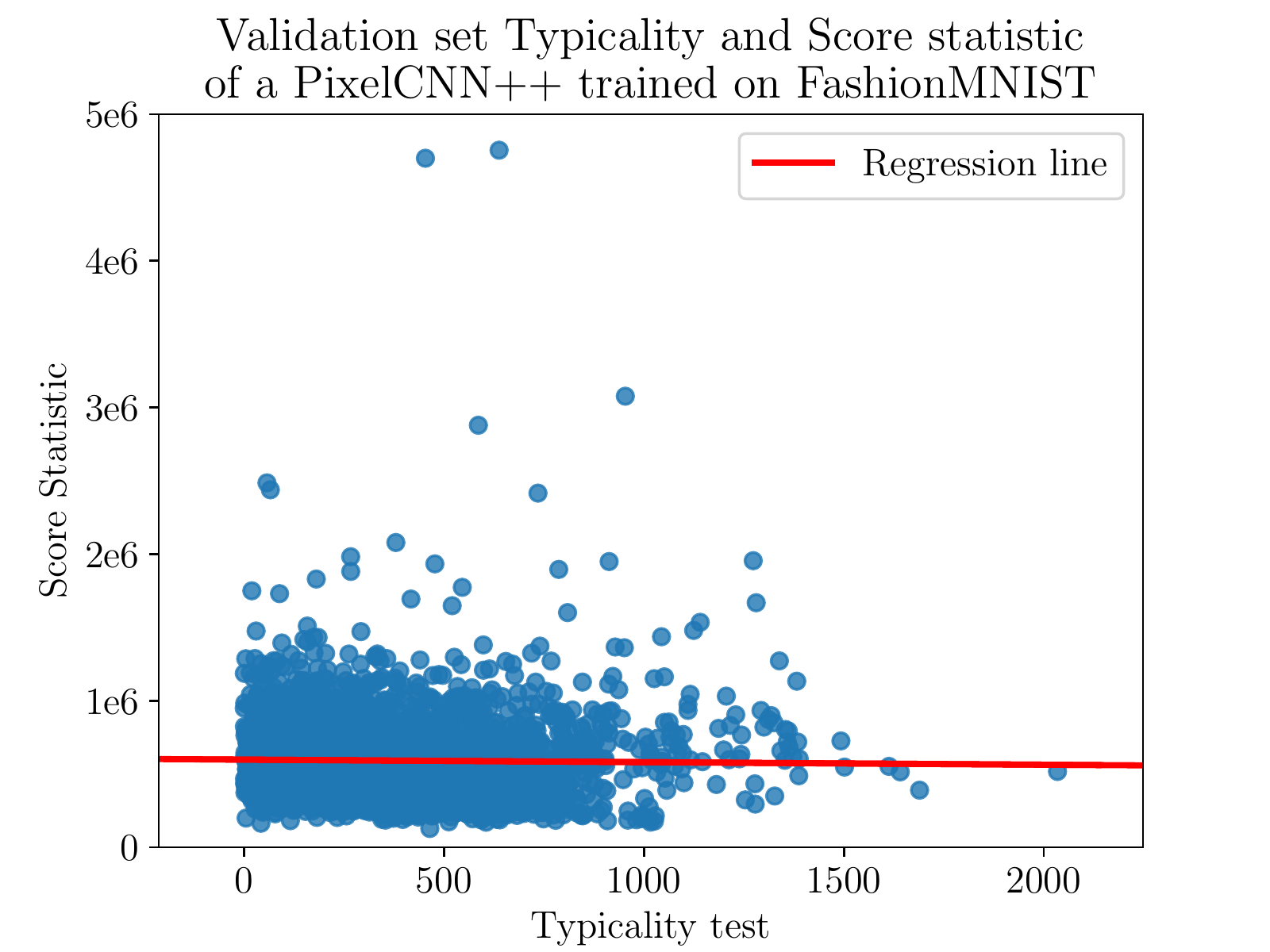}
    \caption{Correlation of Typicality Test and Score Statistic computed on the validation set using a PixelCNN++ trained on FashionMNIST. The correlation coefficient is $-0.014$. This can also be seen by looking at the regression line, which is almost straight. }
    \label{fig:corralation}
\end{figure}

To test if the typicality test and the score statistic are uncorrelated, we plot the two scores computed on the validation set. As can be seen from figure \ref{fig:corralation}, we have that the two measures are not correlated as it is also highlight by the correlation coefficient.

\subsection{Harmonic Mean}
\label{appendix:harmonic}
In the paper we mentioned that another way to combine $p$-values from different test statistics is the Harmonic mean \citep{wilson2019harmonic}. This is defined as:
\begin{equation}
    \mathring{p} = \frac{\sum_{i=1}^{k} w_i}{\sum_{i=1}^{k} w_i / p_i},
\end{equation}
where $w_1,\dots,w_k$ are weights that sum up to 1. In our setting, we considered equal weights, i.e.\@ $w_i = 1 / k$. Therefore, if we simply consider two test statistics $T_1$ and $T_2$ and corresponding $p$-values $p_1$ and $p_2$, the harmonic mean $p$-values becomes:
\begin{equation}
    \mathring{p} = \frac{2 p_1 p_2}{p_1 + p_2}.
\end{equation}

As expected, this combination should work better when the statistics that we are combining are somewhat correlated. Indeed, since in our setting we have that the typicality and the score statistic are independent, we would expect this to work worse than the Fisher's combination. This is confirmed by table~\ref{tab:all_combinations}, where we are reporting the results when combining the two statistics using the three different ways we analyzed.

\subsection{Results considering maximum-mean-discrepancy}
In Section~\ref{sec:section4}, we discussed the relationship between the maximum-mean-discrepancy with a Fisher kernel and the score statistic and the gradient norm, which depends on the choice of approximation of the Fisher information matrix we use. In table~\ref{tab:mmd} we reported also the AUROC scores for the MMD with Fisher kernel considering both the diagonal approximation of the FIM (called \emph{MMD Diagonal} in the table) and the FIM being the identity matrix (called \emph{MMD Identity}). As expected, we have that the AUROC of the MMD with the diagonal approximated FIM is pretty close to the AUROC we obtained by using the score statistic. Likewise, we have that the AUROC of MMD with the identity matrix as FIM is close to the gradient norm when we trained on FashionMNIST and CIFAR10. 

So, why did we decide to use the score statistic instead of the MMD with Fisher kernel and diagonal approximation of the FIM? The main reason is Occam's razor. If we have two things that work equally well, we should keep the simplest one. In our case, we have that for computing the MMD with the Fisher kernel, we need to compute both the average gradient and the FIM using the training set. For the score statistic, instead, we just need the FIM. In addition to that, from all our experiments (see table~\ref{tab:mmd} and table~\ref{tab:mmd_pair2}) we do not have any evidence for one statistic working better than the other, because they are always pretty close to each other.

\begin{table*}[tb]
    \caption{AUROC$\uparrow$ for single-sample OOD detection. In this table we consider all the different single statistics we mentioned in the paper. One can notice that MMD Diagonal is pretty close to the score statistic and the MMD Identity is close to the gradient norm, as expected (see Section 4.1 in the paper).}
    \resizebox{\textwidth}{!}{
        \scriptsize
        \begin{tabular}{ccccccc}
            \toprule
            &\multicolumn{6}{c}{\textsc{FashionMNIST (in) / MNIST (out)}}\\
            \cmidrule{2-7}
            & \multicolumn{6}{c}{\textcolor{blue}{\textsc{single statistics}}}\\
            \cmidrule{2-7}
            \textsc{models}  & \textcolor{blue}{\textsc{$\log p(x)$}} & \textcolor{blue}{\textsc{$\|\nabla \log p(x)\|_2$}} & \textcolor{blue}{\textsc{MMD Diagonal}}  & \textcolor{blue}{\textsc{MMD Identity}} &\textcolor{blue}{\textsc{Typicality}} & \textcolor{blue}{\textsc{Score Stat}}  \\
            \midrule
            \textsc{PixelCNN++} (dropout) & 0.0762 & 0.8709 & 0.8903 & 0.8690 & 0.8314 & 0.8822 \\
            \textsc{PixelCNN++} (no dropout)  & 0.1048 &  0.9532 & 0.9393 & 0.9539 & 0.7575 &  0.9381\\
            \textsc{Glow} (RMSProp) &  0.1970 & 0.8904 & 0.9115 &  0.8986 & 0.4807 &  0.9114\\
            \textsc{Glow} (Adam)  &0.1223 & 0.7705 & 0.8540 & 0.7217 & 0.6987 &  0.8745\\
            \textsc{HVAE}  &0.0653 & 0.8714 & 0.9574 & 0.8726 & 0.8336 & 0.9578 \\
            \bottomrule
            & & & & & & \\
            \toprule
            &\multicolumn{6}{c}{\textsc{CIFAR10 (in) / SVHN (out)}}\\
            \cmidrule{2-7}
            & \multicolumn{6}{c}{\textcolor{blue}{\textsc{single statistics}}}\\
            \cmidrule{2-7}
            \textsc{models}  & \textcolor{blue}{\textsc{$\log p(x)$}} & \textcolor{blue}{\textsc{$\|\nabla \log p(x)\|_2$}} & \textcolor{blue}{\textsc{MMD Diagonal}}  & \textcolor{blue}{\textsc{MMD Identity}} &\textcolor{blue}{\textsc{Typicality}} & \textcolor{blue}{\textsc{Score Stat}}  \\
            \midrule
        
            \textsc{PixelCNN++} (model1)  & 0.1553 & 0.8006 & 0.6406 & 0.8126 & 0.6457 & 0.6407 \\
            \textsc{PixelCNN++} (model2) & 0.1567 & 0.7923 & 0.7070 & 0.7955 & 0.6498 &  0.7067\\
            \textsc{Glow} (RMSProp)  & 0.0630 & 0.8585 & 0.7929 & 0.8621 & 0.8651 &  0.7940\\
            \textsc{Glow} (Adam)   & 0.0627 & 0.7844 & 0.7620 &  0.7838 &  0.8624 &  0.7655 \\
            \textsc{HVAE}  & 0.0455 & 0.8041 &  0.7268 & 0.7634 & 0.8845 & 0.7334\\
            \bottomrule
        \end{tabular}
        \label{tab:mmd}
    }
    \vspace*{-0.85\baselineskip}
\end{table*}

\begin{table*}[tb]
    \centering
    \caption{AUROC$\uparrow$ for single-sample OOD detection. Comparison between the three method we mentioned to combine different statistics. Since the typicality and the score statistic are not correlated, we have that the Fisher's method is mostly working better than the other two methods.}
        \scriptsize
        \begin{tabular}{cccc}
            \toprule
            &\multicolumn{3}{c}{\textsc{FashionMNIST (in) / MNIST (out)}}\\
            \cmidrule{2-4}
            & \multicolumn{3}{c}{\textcolor{red}{\textsc{combinations}}}\\
            \cmidrule{2-4}
            \textsc{models}  & \textcolor{red}{\textsc{Fisher's method}} & \textcolor{red}{\textsc{Harmonic mean}} &  \textcolor{red}{\textsc{DoSE$_{\textup{KDE}}$}}  \\
            \midrule
            \textsc{PixelCNN++} (dropout) & 0.9369 & 0.9148 & 0.8822 \\
            \textsc{PixelCNN++} (no dropout)  & 0.9536 & 0.9392 & 0.9382\\
            \textsc{Glow} (RMSProp) &  0.8598 & 0.8853 &  0.8901\\
            \textsc{Glow} (Adam)  & 0.8839 & 0.8632 & 0.8752\\
            \textsc{HVAE}  & 0.9708 &  0.9569 & 0.9630\\
            \bottomrule
            & & &  \\
            \toprule
            & \multicolumn{3}{c}{\textsc{CIFAR10 (in) / SVHN (out)}}\\
            \cmidrule{2-4}
            & \multicolumn{3}{c}{\textcolor{red}{\textsc{combinations}}}\\
            \cmidrule{2-4}
            \textsc{models}  & \textcolor{red}{\textsc{Fisher's method}} & \textcolor{red}{\textsc{Harmonic mean}} &  \textcolor{red}{\textsc{DoSE$_{\textup{KDE}}$}}  \\
            \midrule
        
            \textsc{PixelCNN++} (model1)  & 0.6826 & 0.6667 & 0.6571  \\
            \textsc{PixelCNN++} (model2) & 0.7300 & 0.7105 & 0.7243 \\
            \textsc{Glow} (RMSProp)  &  0.8683 & 0.8551 & 0.8510\\
            \textsc{Glow} (Adam) & 0.8613 & 0.8493 & 0.8588  \\
            \textsc{HVAE}  & 0.8699 &  0.8525 & 0.8245 \\
            \bottomrule
        \end{tabular}
        \label{tab:all_combinations}
    \vspace*{-\baselineskip}
\end{table*}

\subsection{Variability within the same model in different checkpoints}
\label{appendix:checkpoints}
As mentioned in the paper, we noticed that all statistics depend on choices we made about our model and the training procedure, such as deciding between Adam or RMSProp, or between using dropout or not. In addition to that, we find out that they can differ also within the same model at different checkpoints that obtain almost the same log-likelihood. Here we consider two Glow models, one trained with Adam and one using RMSProp on CIFAR10. For both, we consider two checkpoints that achieve the same test log-likelihood. Those trained with Adam get a log-likelihood of $3.63$ bits/dim, while the ones trained with RMSProp get $3.62$ bits/dim. Results are shown in Table~\ref{tab:variability}. It can be noticed, that although the models are similar in terms of test bits/dim the statistics vary a lot, mostly when training with RMSProp.

\begin{table*}[tb]
    \centering
    \caption{AUROC$\uparrow$ for single-sample OOD detection. In this table we are comparing two different Glow models trained on CIFAR10 by considering two different checkpoints with almost the same test log-likelihood. We can see that both statistics vary a bit. }
        \scriptsize
        \begin{tabular}{ccccc}
            \toprule
            &\multicolumn{4}{c}{\textsc{CIFAR10 (in) / SVHN (out)}}\\
            \cmidrule{2-5}
            & \multicolumn{2}{c}{\textcolor{blue}{\textsc{single statistics}}} & \multicolumn{2}{c}{\textcolor{red}{\textsc{combination}}}\\
            \cmidrule(r){2-3}\cmidrule(l){4-5}
            \textsc{models}  &  \textcolor{blue}{\textsc{Typicality}} & \textcolor{blue}{\textsc{Score Stat}} & \textcolor{red}{\textsc{Fisher's method}} & \textcolor{red}{\textsc{DoSE$_{\textup{KDE}}$}} \\
            \midrule
            \textsc{Glow} (RMSProp) \{\emph{check1}\} & 0.8651 &  0.7940  &  0.8683 &  0.8510\\
            \textsc{Glow} (RMSProp) \{\emph{check2}\} & 0.8532 &  0.6894  & 0.8275 & 0.7815\\
            \midrule
            \textsc{Glow} (Adam)  \{\emph{check1}\} & 0.8624 &  0.7655 & 0.8613 & 0.8588\\
            \textsc{Glow} (Adam) \{\emph{check2}\} & 0.8558 & 0.7327 & 0.8402 & 0.8303\\
            \bottomrule
        \end{tabular}
        \label{tab:variability}
    \vspace*{-\baselineskip}
\end{table*}

\subsection{Benjamini-Hochberg procedure when training on CIFAR10}
\label{appendix:BH_cifar}
\begin{figure}[tb]
    \centering
    \includegraphics[scale=0.5]{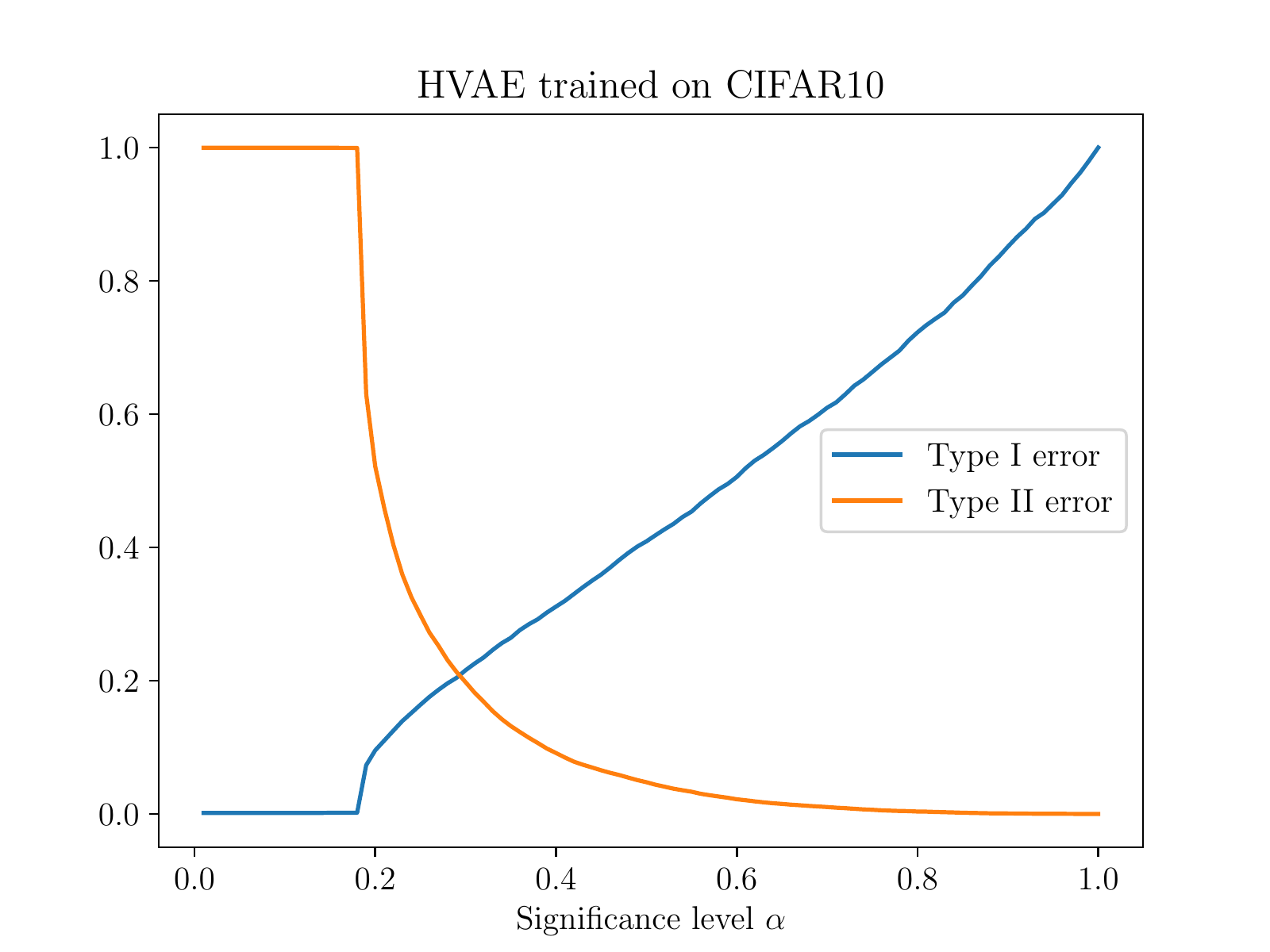}
    \caption{Type I and Type II errors versus the significance level $\alpha$ on the combination values. We can control the FDR only for $\alpha > 0.2$ in this case. For $\alpha >0.2$, since we are using Benjamini-Hochberg procedure, we get that the Type I error stays below identity line.}
    \label{fig:type1_cifar}
     \vspace*{-\baselineskip}
\end{figure}

In the main paper we focused on the Benjamini-Hochberg procedure applied to a model trained on FashionMNIST. Although one should use a False Discovery Rate control procedure when the statistics we are using are strong, for completeness, we will present what happens when we apply the BH procedure on a model trained on CIFAR10. In Fig.~\ref{fig:type1_cifar}, we report the Type I error ratio and the Type II error ratio for different significance levels $\alpha$. We can see that we can actually control the FDR for $\alpha > 0.2$, and for these significance levels we are actually controlling the FDR. What is happening for $\alpha < 0.2$? We have that the procedure is only rejecting 5 hypotheses and all these hypotheses corresponds to in-distribution examples. Therefore, we have that the ratio of Type I error is still low, but we are making a lot of Type II errors because we are accepting all the examples whose hypotheses should be rejected.

\subsection{Results when training on MNIST and SVHN}
\label{appendix:mnist_svhn}
We also evaluated our methods in the two dataset pairs, MNIST against FashionMNIST and SVHN against CIFAR10, that are usually considered easier than the tasks presented in the main paper. For both tasks, we trained two Glow models, one trained with Adam and one trained with RMSProp, one PixelCNN++ trained with dropout and a hierarchical-VAE. Results are reported in table~\ref{tab:results_second}. We can see that almost all the statistics we considered are able to almost perfectly distinguish between the in-distribution test-set and the OOD test-set. However, we can notice that the gradient norm is failing sometimes both when we trained on CIFAR10 and when we trained on FashionMNIST. From table~\ref{tab:mmd_pair2}, instead, it is clear that we need to approximate the diagonal of the Fisher Information Matrix because if we simply consider the identity matrix, this will also fail as the gradient norm is doing.

\begin{table*}[tb]
    \caption{AUROC$\uparrow$ for single-sample OOD detection when training on MNIST and testing again FashionMNIST and when training on SVHN and testing against CIFAR10. As before, Fisher's method is the combination of the typicality test and the test statistic. These are also combined using DoSE.}
    \resizebox{\textwidth}{!}{
        \scriptsize
        \begin{tabular}{ccccccc}
            \toprule
            &\multicolumn{6}{c}{\textsc{MNIST (in) / FashionMNIST (out)}}\\
            \cmidrule{2-7}
            & \multicolumn{4}{c}{\textcolor{blue}{\textsc{single statistics}}} & \multicolumn{2}{c}{\textcolor{red}{\textsc{combination}}}\\
            \cmidrule(r){2-5}\cmidrule(l){6-7}
            \textsc{models}  & \textcolor{blue}{\textsc{$\log p(x)$}} & \textcolor{blue}{\textsc{$\|\nabla \log p(x)\|_2$}} & \textcolor{blue}{\textsc{Typicality}} & \textcolor{blue}{\textsc{Score Stat}} & \textcolor{red}{\textsc{Fisher's method}} & \textcolor{red}{\textsc{DoSE$_{\textup{KDE}}$}} \\
            \midrule
            \textsc{PixelCNN++} (dropout) ($\dagger$)  & 0.9999  & 0.8534 & 0.9996 & 0.9993 & 0.9999  & 0.9999 \\
            \textsc{Glow} (RMSProp) & 0.9997  & 0.9936  & 0.9991 & 0.9936 &  0.9992 & 0.9994\\
            \textsc{Glow} (Adam)  &0.9999 & 0.6506 & 0.9995 &  0.9992 &  0.9998 &  0.9999\\
            \textsc{HVAE}  & 0.9999 & 0.9998 &  0.9997 &  0.9999 &  0.9999  & 0.9999\\
            \bottomrule
            & & & & & & \\
            \toprule
            &\multicolumn{6}{c}{\textsc{SVHN (in) / CIFAR10 (out)}}\\
            \cmidrule{2-7}
            & \multicolumn{4}{c}{\textcolor{blue}{\textsc{single statistics}}} & \multicolumn{2}{c}{\textcolor{red}{\textsc{combination}}}\\
            \cmidrule(r){2-5}\cmidrule(l){6-7}
            \textsc{models}  & \textcolor{blue}{\textsc{$\log p(x)$}} & \textcolor{blue}{\textsc{$\|\nabla \log p(x)\|_2$}} & \textcolor{blue}{\textsc{Typicality}} & \textcolor{blue}{\textsc{Score Stat}} & \textcolor{red}{\textsc{Fisher's method}} & \textcolor{red}{\textsc{DoSE$_{\textup{KDE}}$}} \\
            \midrule
            \textsc{PixelCNN++} (dropout)  & 0.9820 & 0.2670 &  0.9590 &  0.9543 & 0.9914 & 0.9824 \\
            \textsc{Glow} (RMSProp)  &  0.9917  & 0.9180 & 0.9830 & 0.9823 & 0.9913  & 0.9913 \\
            \textsc{Glow} (Adam)   &0.9913 &  0.5658 & 0.9779 & 0.9641 & 0.9883 & 0.9863 \\
            \textsc{HVAE}  & 0.9943 &  0.1011  &  0.9857 &  0.9862 & 0.9934 &  0.9862 \\
            \bottomrule
            ($\dagger$) Trained using 50000 datapoints
        \end{tabular}
        \label{tab:results_second}
    }
\end{table*}

\begin{table*}[tb]
    \caption{AUROC$\uparrow$ for single-sample OOD detection. In this table we consider all the different single statistics we mentioned in the paper but for the models trained on MNIST and SVHN this time. In this case, it is important to notice that the gradient norm and the MMD identity sometimes fail to a different extent.}
    \resizebox{\textwidth}{!}{
        \scriptsize
        \begin{tabular}{ccccccc}
            \toprule
            &\multicolumn{6}{c}{\textsc{MNIST (in) / FashionMNIST (out)}}\\
            \cmidrule{2-7}
            & \multicolumn{6}{c}{\textcolor{blue}{\textsc{single statistics}}}\\
            \cmidrule{2-7}
            \textsc{models}  & \textcolor{blue}{\textsc{$\log p(x)$}} & \textcolor{blue}{\textsc{$\|\nabla \log p(x)\|_2$}} & \textcolor{blue}{\textsc{MMD Diagonal}}  & \textcolor{blue}{\textsc{MMD Identity}} &\textcolor{blue}{\textsc{Typicality}} & \textcolor{blue}{\textsc{Score Stat}}  \\
            \midrule
            \textsc{PixelCNN++} (dropout) ($\dagger$)  & 0.9999  & 0.8534 & 0.9993 & 0.8608 & 0.9996  & 0.9993 \\
            \textsc{Glow} (RMSProp) & 0.9997  & 0.9936  & 0.9942 & 0.6609 &  0.9991 & 0.9936\\
            \textsc{Glow} (Adam)  &0.9999 & 0.6506 & 0.9993 &  0.9124 &  0.9997 &  0.9992\\
            \textsc{HVAE}  & 0.9999 & 0.9998 &  0.9999 &  0.9999 &  0.9999  & 0.9999\\
            \bottomrule
            & & & & & & \\
            \toprule
            &\multicolumn{6}{c}{\textsc{SVHN (in) / CIFAR10 (out)}}\\
            \cmidrule{2-7}
            & \multicolumn{6}{c}{\textcolor{blue}{\textsc{single statistics}}}\\
            \cmidrule{2-7}
            \textsc{models}  & \textcolor{blue}{\textsc{$\log p(x)$}} & \textcolor{blue}{\textsc{$\|\nabla \log p(x)\|_2$}} & \textcolor{blue}{\textsc{MMD Diagonal}}  & \textcolor{blue}{\textsc{MMD Identity}} &\textcolor{blue}{\textsc{Typicality}} & \textcolor{blue}{\textsc{Score Stat}}  \\
            \midrule
            \textsc{PixelCNN++} (dropout)  & 0.9820 & 0.2670 &   0.9543 & 0.3185 & 0.9590 & 0.9543 \\
            \textsc{Glow} (RMSProp)  &  0.9917  & 0.9180 & 0.9824 & 0.9317 & 0.9830  &  0.9823 \\
            \textsc{Glow} (Adam)   &0.9913 &  0.5658 & 0.9653 & 0.7096 & 0.9779 &  0.9641 \\
            \textsc{HVAE}  & 0.9943 &  0.1011  &  0.9865 &  0.4508 & 0.9857 &  0.9862\\
            \bottomrule
            ($\dagger$) Trained using 50000 datapoints
        \end{tabular}
        \label{tab:mmd_pair2}
    }
\end{table*}

\subsection{Application of our method to Gaussian Mixture Model and Probabilistic PCA}
\label{appendix:gmm_ppca}
Since the method we propose is model-agnostic, we show that it can be used for out-of-distribution detection also using two simple generative models, Gaussian Mixture Model (GMM) and Probabilistic PCA (PPCA). We consider the two pairs of datasets as before, i.e.\@ FashionMNIST vs MNIST and CIFAR10 vs SVHN. Results can be seen in Table~\ref{tab:gmm_results} and Table~\ref{tab:ppca_results}. For both GMM and PPCA trained on FashionMNIST the likelihood can be used to perform OOD detection. Indeed, in this setting, they are not assigning higher likelihood to OOD data as it is the case for DGMs. This happens instead when we fit these models on CIFAR10. However, this behaviour can be due to the fact that they are really poor generative models for this dataset. It is also surprising that when training on CIFAR10 the score statistic is failing in both models. We think that this is also due to the fact that both the GMM and the PPCA are far from being good generative models for this dataset.

\begin{table*}[tb]
    \caption{AUROC$\uparrow$ for single-sample OOD detection using a Gaussian mixture model (GMM). For Fisher's method we mean the combination of the typicality test and the test statistic. These are also combined using DoSE.}
    \resizebox{\textwidth}{!}{
        \scriptsize
        \begin{tabular}{ccccccc}
            \toprule
            &\multicolumn{6}{c}{\textsc{FashionMNIST (in) / MNIST (out)}}\\
            \cmidrule{2-7}
            & \multicolumn{4}{c}{\textcolor{blue}{\textsc{single statistics}}} & \multicolumn{2}{c}{\textcolor{red}{\textsc{combination}}}\\
            \cmidrule(r){2-5}\cmidrule(l){6-7}
            \textsc{components}  & \textcolor{blue}{\textsc{$\log p(x)$}} & \textcolor{blue}{\textsc{$\|\nabla \log p(x)\|_2$}} & \textcolor{blue}{\textsc{Typicality}} & \textcolor{blue}{\textsc{Score Stat}} & \textcolor{red}{\textsc{Fisher's method}} & \textcolor{red}{\textsc{DoSE$_{\textup{KDE}}$}} \\
            \midrule
            \textsc{50}  & 0.6627 & 0.5514  & 0.5196 & 0.8777  & 0.7689 & 0.8152 \\
            \textsc{100}   &  0.6872 & 0.5509 & 0.5575 & 0.8742 & 0.7965 & 0.7989 \\
            \bottomrule
            &  &  &   &   &  & \\
            \toprule
            &\multicolumn{6}{c}{\textsc{CIFAR10 (in) / SVHN (out)}}\\
            \cmidrule{2-7}
            & \multicolumn{4}{c}{\textcolor{blue}{\textsc{single statistics}}} & \multicolumn{2}{c}{\textcolor{red}{\textsc{combination}}}\\
            \cmidrule(r){2-5}\cmidrule(l){6-7}
            \textsc{components}  & \textcolor{blue}{\textsc{$\log p(x)$}} & \textcolor{blue}{\textsc{$\|\nabla \log p(x)\|_2$}} & \textcolor{blue}{\textsc{Typicality}} & \textcolor{blue}{\textsc{Score Stat}} & \textcolor{red}{\textsc{Fisher's method}} & \textcolor{red}{\textsc{DoSE$_{\textup{KDE}}$}} \\
            \midrule
            \textsc{50}   & 0.2335  &   0.6087 & 0.6759 & 0.3512 &  0.6098 & 0.6569 \\
            \textsc{100} & 0.2372 &  0.6136 & 0.6714 & 0.3294  &  0.5898  & 0.6573 \\
            \bottomrule
        \end{tabular}
        \label{tab:gmm_results}
    }
    \vspace*{-\baselineskip}
\end{table*}

\begin{table*}[tb]
    \caption{AUROC$\uparrow$ for single-sample OOD detection using a Probabilistic PCA. For Fisher's method we mean the combination of the typicality test and the test statistic. These are also combined using DoSE.}
    \resizebox{\textwidth}{!}{
        \scriptsize
        \begin{tabular}{ccccccc}
            \toprule
            &\multicolumn{6}{c}{\textsc{FashionMNIST (in) / MNIST (out)}}\\
            \cmidrule{2-7}
            & \multicolumn{4}{c}{\textcolor{blue}{\textsc{single statistics}}} & \multicolumn{2}{c}{\textcolor{red}{\textsc{combination}}}\\
            \cmidrule(r){2-5}\cmidrule(l){6-7}
            \textsc{components}  & \textcolor{blue}{\textsc{$\log p(x)$}} & \textcolor{blue}{\textsc{$\|\nabla \log p(x)\|_2$}} & \textcolor{blue}{\textsc{Typicality}} & \textcolor{blue}{\textsc{Score Stat}} & \textcolor{red}{\textsc{Fisher's method}} & \textcolor{red}{\textsc{DoSE$_{\textup{KDE}}$}} \\
            \midrule
            \textsc{50}  & 0.9727 & 0.9637 & 0.9587 & 0.9505  & 0.9635 & 0.9610 \\
            \textsc{100}   & 0.9557 &  0.9715 & 0.9309 & 0.9626 & 0.9566 & 0.9585\\
            \bottomrule
            &  &  &   &   &  & \\
            \toprule
            &\multicolumn{6}{c}{\textsc{CIFAR10 (in) / SVHN (out)}}\\
            \cmidrule{2-7}
            & \multicolumn{4}{c}{\textcolor{blue}{\textsc{single statistics}}} & \multicolumn{2}{c}{\textcolor{red}{\textsc{combination}}}\\
            \cmidrule(r){2-5}\cmidrule(l){6-7}
            \textsc{components}  & \textcolor{blue}{\textsc{$\log p(x)$}} & \textcolor{blue}{\textsc{$\|\nabla \log p(x)\|_2$}} & \textcolor{blue}{\textsc{Typicality}} & \textcolor{blue}{\textsc{Score Stat}} & \textcolor{red}{\textsc{Fisher's method}} & \textcolor{red}{\textsc{DoSE$_{\textup{KDE}}$}} \\
            \midrule
            \textsc{50}  & 0.0770 & 0.1494  &  0.8468 & 0.1308 & 0.7568 & 0.8210 \\
            \textsc{100}   &  0.0357 &  0.0778 & 0.8944 & 0.0755  &  0.7966 &  0.8830 \\
            \bottomrule
        \end{tabular}
        \label{tab:ppca_results}
    }
\end{table*}

\subsection{More in depth analysis of the variability of the results for different HVAE}
As we have pointed out before, test statistics and consequentially out-of-distribution performances can vary between the same model trained several times on the same dataset. To test the variability of the results shown in the main paper, we trained five different hierarchical VAEs and compute mean and standard deviations of the final AUROC scores. All models have the same architecture and were trained with the same procedure. Results can be found in Table~\ref{table:consistency_results}. For the models trained on CIFAR10, most of the variability in terms of performance is due to the score statistic, which has the highest standard deviation. When training on FashionMNIST, instead, it seems that the typicality performance is the one varying the most between the five models.

\begin{table*}[tb]
    \caption{Mean and standard deviation of the performance in terms of AUROC of our method. Quantities are computed by taking the performance of 5 different trained HVAEs both trained on CIFAR10 and FashionMNIST.}
        \centering
        \scriptsize
        \begin{tabular}{cccccc}
            \toprule
            \textsc{$D_{\text{out}}$}  & \textcolor{blue}{\textsc{$\log p(x)$}} & \textcolor{blue}{\textsc{Typicality}} & \textcolor{blue}{\textsc{Score Stat}} & \textcolor{red}{\textsc{Fisher's method}} & \textcolor{red}{\textsc{DoSE$_{\textup{KDE}}$}}  \\
            \midrule
            &\multicolumn{5}{c}{\textsc{HVAE trained on CIFAR10}} \\
            \cmidrule{2-6}
            \textsc{SVHN} & 0.0631 (0.0008) & 0.8711 (0.0028)  & 0.7808 (0.0255) & 0.8844 (0.0140)  & 0.8519 (0.0194)\\
            \textsc{CIFAR100} & 0.5349 (0.0007) & 0.5496 (0.0003)  & 0.5857 (0.0042) & 0.5924 (0.0029)  & 0.5985 (0.0028)\\
            \textsc{CelebA} & 0.9004 (0.0035) & 0.8203 (0.0046) & 0.7565 (0.0369) & 0.8505 (0.0138) & 0.8247 (0.0228)\\
            \cmidrule{2-6}
            &\multicolumn{5}{c}{\textsc{HVAE trained on FashionMNIST}} \\
            \cmidrule{2-6}%
            \textsc{MNIST} & 0.2487 (0.0152) & 0.5064 (0.0245) & 0.9532 (0.0084) & 0.9220 (0.01491) & 0.9377 (0.0126)\\
            \bottomrule%
    \end{tabular}
    \label{table:consistency_results}
\end{table*}

\section{Yes, we should talk about CelebA}
\label{appendix:celeba}
Out-of-distribution detection performance is not only influenced by the model architecture or the training process. Indeed, transformations applied to the input data play an important role by transforming a difficult task into an easier problem where the likelihood can detect OOD data. By looking at the different results for Glow trained on CIFAR10 and tested on CelebA shown in \cite{hendrycks2018deep}, \cite{kirichenko2020normalizing}, \cite{morningstar2021density}, and \cite{ijcai2021-292} we can see that the AUROC scores obtain by the plain log-likelihood are pretty different. In \cite{hendrycks2018deep} and  \cite{kirichenko2020normalizing} the log-likelihood gets a poor performance, confirming that CIFAR10-CelebA is a challenging pair for DGMs, while in \cite{morningstar2021density} the likelihood is able to distinguish OOD data. While the main reason for these different results can be due to model implementation and training procedure, we decided to investigate how different transformations can influence OOD detection. Indeed, CelebA examples originally have a shape of $(218, 178, 3)$ and to transform them into $(32, 32, 3)$-shaped images, as CIFAR10, we have to resize them and then crop their center. The resize function is performing an interpolation, therefore we analyze how different interpolation strategies influence the OOD task. 

\begin{figure}[t]
    \centering
    \includegraphics[scale=0.5]{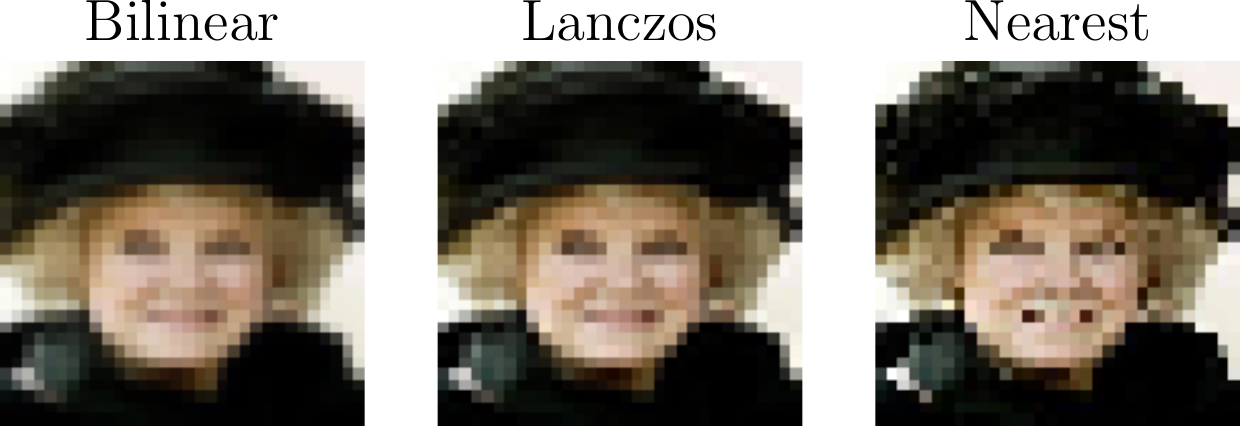}
    \caption{Comparison of different interpolation methods for CelebA dataset.}
    \label{fig:celeba}
     \vspace*{-\baselineskip}
\end{figure}

We considered three different interpolations: bilinear (default in PyTorch), Lanczos, and nearest. As can be seen from Fig.~\ref{fig:celeba}, these transformations mostly affect the sharpness of the images. In Table~\ref{table:celeba} we show how the OOD performance changes for our considered models when testing on CelebA where we applied different interpolations. We can notice that when using the bilinear interpolation we get results that are pretty similar to \cite{hendrycks2018deep}, \cite{kirichenko2020normalizing}, and \cite{ijcai2021-292} in terms of likelihood OOD performance. When using the nearest interpolation, instead, we get results that are closer to \cite{morningstar2021density}. 

In conclusion, with these experiments, we wanted to highlight the importance of reporting the preprocessing steps used in loading CelebA in order to be able to make a fair comparison with the other proposed methods in the literature.

\begin{table*}[tb]
    \caption{AUROC$\uparrow$ for single-sample OOD detection for CIFAR10 vs CelebA considering all the three interpolations when using CelebA.}
    \resizebox{\textwidth}{!}{
        \scriptsize
        \begin{tabular}{ccccccc}
            \toprule
            &\multicolumn{6}{c}{\textsc{CIFAR10 (in) / CelebA (out) ($\dagger$)}}\\
            \cmidrule{2-7}
            & \multicolumn{4}{c}{\textcolor{blue}{\textsc{single statistics}}} & \multicolumn{2}{c}{\textcolor{red}{\textsc{combination}}}\\
            \cmidrule(r){2-5}\cmidrule(l){6-7}
            \textsc{models}  & \textcolor{blue}{\textsc{$\log p(x)$}} & \textcolor{blue}{\textsc{$\|\nabla \log p(x)\|_2$}} & \textcolor{blue}{\textsc{Typicality}} & \textcolor{blue}{\textsc{Score Stat}} & \textcolor{red}{\textsc{Fisher's method}} & \textcolor{red}{\textsc{DoSE$_{\textup{KDE}}$}} \\
            \midrule
            \textsc{PixelCNN++} (model1)  &  0.7027 & 0.5856 & 0.5581  & 0.7001 & 0.6450 & 0.6931 \\
            \textsc{PixelCNN++} (model2) & 0.7034  & 0.4298 & 0.5554 & 0.7505 & 0.6879 & 0.7430\\
            \textsc{Glow} (RMSProp)  & 0.5337  & 0.5616 &  0.3926 & 0.6561  & 0.5400 & 0.5866 \\
            \textsc{Glow} (Adam)   & 0.5308  & 0.5820 & 0.3914 & 0.5850  & 0.4818 & 0.5212 \\
            \textsc{HVAE}  &  0.5643 & 0.5214  & 0.4011 & 0.6712 & 0.5483 & 0.5987 \\
            \bottomrule
            & & & & & & \\
            \toprule
            &\multicolumn{6}{c}{\textsc{CIFAR10 (in) / CelebA (out) ($\nparallel$)}}\\
            \cmidrule{2-7}
            & \multicolumn{4}{c}{\textcolor{blue}{\textsc{single statistics}}} & \multicolumn{2}{c}{\textcolor{red}{\textsc{combination}}}\\
            \cmidrule(r){2-5}\cmidrule(l){6-7}
            \textsc{models}  & \textcolor{blue}{\textsc{$\log p(x)$}} & \textcolor{blue}{\textsc{$\|\nabla \log p(x)\|_2$}} & \textcolor{blue}{\textsc{Typicality}} & \textcolor{blue}{\textsc{Score Stat}} & \textcolor{red}{\textsc{Fisher's method}} & \textcolor{red}{\textsc{DoSE$_{\textup{KDE}}$}} \\
            \midrule
            \textsc{PixelCNN++} (model1)  &  0.8284  & 0.5035 & 0.7399 & 0.6714  & 0.7477 & 0.7123 \\
            \textsc{PixelCNN++} (model2) & 0.8284 & 0.3530 &   0.7370 & 0.70088 & 0.7631 & 0.7446\\
            \textsc{Glow} (RMSProp)  & 0.7556 & 0.4427 &  0.6222 & 0.7865  & 0.7423 & 0.7632 \\
            \textsc{Glow} (Adam)   & 0.7499 & 0.4800 &  0.6177 &  0.6442  &  0.6460 &  0.6467 \\
            \textsc{HVAE}  & 0.7561 &   0.4097 & 0.6051 &  0.6779 &  0.6775 &  0.6772  \\
            \bottomrule
             & & & & & & \\
            \toprule
            &\multicolumn{6}{c}{\textsc{CIFAR10 (in) / CelebA (out) ($\ddagger$)}}\\
            \cmidrule{2-7}
            & \multicolumn{4}{c}{\textcolor{blue}{\textsc{single statistics}}} & \multicolumn{2}{c}{\textcolor{red}{\textsc{combination}}}\\
            \cmidrule(r){2-5}\cmidrule(l){6-7}
            \textsc{models}  & \textcolor{blue}{\textsc{$\log p(x)$}} & \textcolor{blue}{\textsc{$\|\nabla \log p(x)\|_2$}} & \textcolor{blue}{\textsc{Typicality}} & \textcolor{blue}{\textsc{Score Stat}} & \textcolor{red}{\textsc{Fisher's method}} & \textcolor{red}{\textsc{DoSE$_{\textup{KDE}}$}} \\
            \midrule
            \textsc{PixelCNN++} (model1)  & 0.9270  & 0.4196 & 0.8902 &  0.8320 & 0.9287 &  0.8908 \\
            \textsc{PixelCNN++} (model2) & 0.9270 & 0.3065 & 0.8886 & 0.8448 & 0.9339 & 0.9236 \\
            \textsc{Glow} (RMSProp)  & 0.9364 & 0.5345 &  0.8880 & 0.9286  & 0.9390 & 0.9423 \\
            \textsc{Glow} (Adam)   &  0.9322 & 0.5957 & 0.8829  &  0.8350  & 0.9017 &  0.8933 \\
            \textsc{HVAE}  & 0.8964 & 0.3515  & 0.8158 & 0.7952 &  0.8620 &   0.8455 \\
            \bottomrule
            ($\dagger$) Bilinear interpolation \\
            ($\nparallel$) Lanczos interpolation \\
            ($\ddagger$) Nearest interpolation\\
        \end{tabular}
        \label{table:celeba}
    }
    \vspace*{-\baselineskip}
\end{table*}

\section{Comparison with the original DoSE statistics}
As the last experiment, we study how our proposed method with our model agnostic statistic performs against DoSE using the original statistics proposed in \cite{morningstar2021density}. For the VAEs model, they suggested to use the following 5 statistics: the posterior/prior cross-entropy $\text{H}[q_{\phi}( \textbf{z}\mid \textbf{x}), p(\textbf{z})]$, the posterior entropy $\text{H}[q_{\phi}(\textbf{z}\mid \textbf{x})]$, the posterior/prior KL divergence $\text{D}_{\text{KL}}[q_{\phi}(\textbf{z}\mid \textbf{x}) \mid \mid p(\textbf{z})]$, the posterior expected log-likelihood $\mathbb{E}_{q_{\phi}(\textbf{z}\mid \textbf{x})}[\log q_{\phi}(\textbf{z}\mid \mathbf{x})]$, and the log-likelihood $\log \mathbb{E}_{q_{\phi}(\textbf{z}\mid \textbf{x})} \left[ \frac{p_{\theta}(\textbf{x}, \textbf{z})}{q_{\phi}(\textbf{z}\mid \textbf{x})}\right]$. For DoSE on Glow, instead, they considered three metrics: the log-likelihood $p_{X}(\textbf{x}\mid \theta_n)$ and its two components, i.e.\@ the log-probability of the latent variable $p_{Z}(\textbf{z}\mid \textbf{x}, \theta_n)$ and the log-determinant of the Jacobian $\log | \text{J}_{f}(\textbf{x})|$. 

In this setting, since DoSE is using statistics that are HVAE and Glow specific, it is not model agnostic anymore. Indeed, we cannot use those statistics also for a PixelCNN++ for example or any other DGM. We want also to highlight that the models used in \cite{morningstar2021density} are a bit different from the ones used in this work. For example, they are considering a beta-VAE with only one stochastic layer, while in our case we used a HVAE with 3-stochastic layers.

\begin{table}%
    \vspace{-1.3em}
    \caption{Comparison between our method and DoSE using the original statistics. In these experiments we considered only Glow trained with Adam.}
        \centering
        \scriptsize
        \begin{tabular}{ccc}
            \toprule
            \textsc{$D_{\text{out}}$}  & \textcolor{red}{\textsc{Our method}} & \textcolor{red}{\textsc{DoSE$_{\textup{orig}}$}} \\
            \midrule
            &\multicolumn{2}{c}{\textsc{GLOW trained on CIFAR10}} \\
            \cmidrule{2-3}
            \textsc{SVHN} & \textbf{0.8613}  &  0.7819  \\
            \textsc{CIFAR100} & \textbf{0.5775} & 0.5700 \\
            \textsc{CelebA} & 0.9017 & \textbf{0.9663}  \\
            \cmidrule{2-3}
            &\multicolumn{2}{c}{\textsc{GLOW trained on FashionMNIST}} \\
            \cmidrule{2-3}%
            \textsc{MNIST} & 0.8839 & \textbf{0.9568}   \\
            \midrule%
            &\multicolumn{2}{c}{\textsc{HVAE trained on FashionMNIST}} \\
            \cmidrule{2-3}%
            \textsc{MNIST} & 0.9383 & \textbf{0.9762} \\
            &\multicolumn{2}{c}{\textsc{HVAE trained on CIFAR10}} \\
            \cmidrule{2-3}%
            \textsc{SVHN} & 0.8605 & \textbf{0.8823}  \\
            \textsc{CIFAR100} & \textbf{0.5888} & 0.5608  \\
            \textsc{CelebA} & \textbf{0.8620} & 0.8203  \\
            \bottomrule
    \end{tabular}
    \label{table:original_dose}
    \vspace*{-\baselineskip}
\end{table}

\section{Algorithmic implementation}
A pseudocode describing step-by-step how to implement our method is given in Algorithm~\ref{algo_description}.

\begin{algorithm}[ht]
\begin{algorithmic}
\STATE \textbf{Input}: Training data $\textbf{X} = (x_1,\ldots,x_m)^T$, validation data $\textbf{X}'$, trained model $p_{\theta}(x)$.\\
\textbf{} \\
\textit{Approximation of the diagonal of the Fisher Information Matrix $I(\theta)$ and average log-likelihood $(1/m)\log p_{\theta}(x_1,\ldots,x_m)$, indicated by $L(\theta)$. We do it in an online fashion.} \\
\textbf{Initialize}  ${I(\theta)} = {0}$ and $L(\theta)=0$ \\
\textbf{For all} $i \in \{1,\ldots ,m\}$: \\
\hspace{1.5em} \textbf{Compute} $\log p_{\theta} ({x}_i)$ \\
\hspace{1.5em} \textbf{Compute} $\nabla_{{\theta}}\log p({x}_i \mid \theta)$ \\
\hspace{1.5em} \textbf{Set}  ${I(\theta)} = \frac{1}{i+1} \cdot (i \cdot {I(\theta)} + (\nabla_{{\theta}}\log p_{\theta}({x}_i))^2$) \\
\hspace{1.5em} \textbf{Set}  $L(\theta) = \frac{1}{i+1} \cdot (i \cdot L(\theta) + \log p_{\theta}({x}_i)$) \\
\textbf{} \\
\textit{Estimation of distributions over the test statistics} \\
\textbf{Sample} $S$ $M'$-sized datasets from $\textbf{X}'$ using bootstrap resampling. \\
\textit{(For single-sample OOD we just cycle through each example, see Sec.~\ref{sec:combination_p_values})}\\
\textbf{Initialize}   $T^{\text{typicality}} = [\hspace{0.5mm}]$ and $T^{\text{score}} = [\hspace{0.5mm}]$ \\
\textbf{For every} bootstrapped dataset $\textbf{X}_s' = (x_1, \dots, x_{M'})^T$: \\
\hspace{1.5em} \textbf{Compute} $\frac{1}{m'} \sum_{m'=1}^{M'}\log p_{\theta}({x}_{m'})$ \\
\hspace{1.5em} \textbf{Compute} $\frac{1}{m'} \sum_{m'=1}^{M'}\nabla_{{\theta}}\log p_{\theta}({x}_{m'})$ \\
\hspace{1.5em} \textbf{Compute} MMD Typicality for ${x}_{m'}$ by $\left\| \frac{1}{m'} \sum_{m'=1}^{M'}\log p_\theta ({x}_{m'}) - L(\theta) \right\|_2$ and add it to $T^{\text{typicality}}$\\
\hspace{1.5em} \textbf{Compute} Score Statistic for ${x}_{m'}$ by $\left\|{I(\theta)}^{-1/2} \frac{1}{m'} \sum_{m'=1}^{M'}\nabla \log p_\theta ({x}_{m'})\right\|_2$ and add it to $T^{\text{score}}$\\
\textbf{Return} Two vectors of size $S$ containing the two statistics for $T^{\text{typicality}}$ and $T^{\text{score}}$  \\
\textbf{} \\
\textbf{Compute} $\hat{F}^{\text{typicality}}$ and $\hat{F}^{\text{score}}$, the two empirical CDFs, from $T^{\text{typicality}}$ and $T^{\text{score}}$. For example, we used \texttt{statsmodels} library \citep{seabold2010statsmodels}. \\
\textbf{} \\  
\textbf{Given} a test set $\tilde{x}_1, \dots, \tilde{x}_n$:\\
\textit{($n=1$ corresponds to perform single-sample OOD detection)} \\ 
\hspace{1.5em} \textbf{Compute} $\frac{1}{n} \sum_{i=1}^{n}\log p_{\theta}(\tilde{{x}}_i)$ and $\frac{1}{n} \sum_{i=1}^{n}\nabla_{{\theta}}\log p_{\theta}(\tilde{{x}}_i)$ \\
\hspace{1.5em} \textbf{Compute} MMD Typicality $\tilde{t}$ and Score Statistic $\tilde{s}$ \\
\hspace{1.5em} \textbf{Compute} $p$-values $p_{T} = 1 - \hat{F}^{\text{typicality}}(\hat{t})$ and $p_{S} = 1 - \hat{F}^{\text{score}}(\tilde{s})$ \\
\hspace{1.5em} \textbf{Combine} the two $p$-values using Fisher's method Eq.~\ref{eq:Fisher_method}
\end{algorithmic}
\caption{Computing $p$-values for OOD detection using a trained generative model.}
\label{algo_description}
\end{algorithm}

\end{document}